\documentclass[preprint,12pt,authoryear]{elsarticle}
\usepackage{tabularx}
\usepackage{graphicx}
\usepackage{adjustbox}
\usepackage[section]{placeins}
\usepackage{float}
\usepackage{enumitem}
\usepackage{algorithm}
\usepackage{dirtytalk}

\usepackage{amssymb}
\usepackage{subfigure}
\usepackage{array}

\usepackage{algpseudocode}

\begin{document}

\begin{frontmatter}



\title{Skin cancer diagnosis using NIR spectroscopy data of skin lesions \textit{in vivo} using machine learning algorithms}

\author[1,8]{Flavio P. Loss}
\author[4,8]{Pedro H. da Cunha}
\author[1,2,8]{Matheus B. Rocha}
\author[3,8]{Madson Poltronieri Zanoni}
\author[1,2,8]{Leandro M. de Lima}
\author[6]{Isadora Tavares Nascimento}
\author[6]{Isabella Rezende}
\author[6,9]{Tania R. P. Canuto}
\author[6,9]{Luciana de Paula Vieira}
\author[6,9]{Renan Rossoni}
\author[7,8]{Maria C. S. Santos}
\author[5,6,8]{Patricia Lyra Frasson}
\author[4,8,10]{Wanderson Romão} 
\author[4,8]{Paulo R. Filgueiras}

% \author[4,8]{Francine sobrenome }
% \author[4,8]{Matheus sobrenome }
% \author[3,8]{Madison sobrenome }

\author[1,2,8]{Renato A. Krohling}
\ead{krohling.renato@gmail.com corresponding author}

\address[1]{Labcin - Nature Inspired Computing Lab.}
\address[2]{PPGI - Graduate Program in Computer Science}
\address[3]{Department of Chemistry}
\address[4]{PPGQ - Graduate Program in Chemistry}
\address[5]{Department of Specialized Medicine}
\address[6]{Dermatological Assistance Program (PAD)}
\address[7]{Pathological Anatomy Unit of the University Hospital Cassiano Antônio Moraes (HUCAM)}
\address[8]{Federal University of Esp\'irito Santo  (UFES), Vit\'oria, ES, 29075-910, Brazil}
\address[9]{Secretary of Health of the Esp\'irito Santo State, Vitória, Brazil}
\address[10]{Federal Institute of Esp\'irito Santo (IFES), Vit\'oria, ES, 29040-780, Brazil}

\begin{abstract}
Skin lesions are classified in benign or malignant. Among the malignant, melanoma is a very aggressive cancer and the major cause of deaths. So, early diagnosis of skin cancer is very desired. In the last few years, there is a growing interest in computer aided diagnostic (CAD) using most image and clinical data of the lesion. Although we have seen an increasing progress in CAD of skin lesions, these sources of information present limitations due to their inability to provide information of the molecular structure of the lesion. NIR spectroscopy may provide an alternative source of information to automated CAD of skin lesions. The most commonly used  techniques and classification algorithms used in spectroscopy are Principal Component Analysis (PCA), Partial Least Squares - Discriminant Analysis (PLS-DA), and Support Vector Machines (SVM). Nonetheless, there is a growing interest in applying the modern techniques of machine and deep learning (MDL) to spectroscopy. One of the main limitations to apply MDL to spectroscopy is the lack of public datasets. Since there is no public dataset of NIR spectral data to skin lesions, as far as we know, an effort has been made and a new dataset named NIR-SC-UFES, has been collected, annotated and analyzed generating the gold-standard for classification of NIR spectral data to skin cancer. Next, the machine learning algorithms XGBoost, CatBoost, LightGBM, 1D-convolutional neural network (1D-CNN) and standard algorithms as SVM and PLS-DA were investigated to classify cancer and non-cancer skin lesions. Experimental results indicate the best performance obtained by LightGBM with pre-processing using standard normal variate  (SNV), feature extraction and data augmentation with Generative Adversarial Networks (GAN)  providing values of 0.839 for balanced accuracy, 0.851 for recall, 0.852 for precision, and 0.850 for F-score. The obtained results indicate the first steps in CAD of skin lesions aiming the automated triage of patients with skin lesions \textit{in vivo} using NIR spectral data.
\end{abstract}


%%\item We provide a new dataset composed of Near-Infrared spectral samples collected \textit{in vivo} of the most common skin lesions, named NIR-SC-UFES, not yet available in the literature;
%%\item We perform extensive experiments with the machine learning algorithms XGBoost, CatBoost, Light GBM and 1D-CNN to classify the NIR spectral dataset;
%%\item We investigate the influence of pre-processing and feature extraction as well as data augmentation techniques to tackle small datasets.


\end{frontmatter}


\section{Introduction}
\label{sec:intro}

%%       \citet{<label>} ==> Jones et al. (2015)
%%       \citep{<label>} ==> (Jones et al., 2015)

In recent years, machine and deep learning algorithms have provided promising results in several areas of knowledge that were previously dominated by statistical or mathematical approaches, indicating also potential to spectral data \citep{carlosa}. However, machine learning algorithms, and especially deep learning algorithms, need a significant amount of data to perform well \citep{Ng2020}.

%%%%%%%%%%%%
One of the first works reporting the application of spectroscopy in the detection of skin cancer is due to \citet{McIntosh2002}. They used Linear Discriminant Analysis (LDA) to classify skin lesions with promising results. Next, \citet{Gniadecka2004} used artificial neural networks to classify spectral data  skin lesions obtaining satisfactory results.

Convolutional Neural Networks (CNN) have also been applied to achieve better spectral data classification results, mainly due to the pre-processing and feature extraction capability provided by the convolutional layers \citep{Liu2017, Malek2017}. \citet{Morais2020} and \citet{Zeng_2021} tackle the spectral data classification problem with the K-Nearest Neighbors (KNN) algorithm.

Data related to infrared spectroscopy of skin lesions present characteristics such as the presence of high-dimensional data, class imbalance and low number of samples, requiring yet additional research efforts to ensure better classification performance.

\citet{Arajo2021} presents an algorithm for extracting pre-established statistical features of Raman spectral data of skin lesions and they applied the Light Gradient Boosting Machine (LightGBM) \citep{Ke2017} as a classification algorithm with promising results to discriminate between nevus and melanoma.

Several advanced gradient boosting machine (GBM) algorithms have been proposed in recent years, such as, the eXtreme Gradient Boosting (XGBoost) \citep{Chen2016}, Categorical Boosting (CatBoost) \citep{Prokhorenkova2018}, and the Light Gradient Boosting Machine (LightGBM). These advanced GBM algorithms have provided state of the art results in classification problems.

This paper aims to investigate the performance of machine and deep learning algorithms for classification of spectral data of the main skin lesions. Since there is no public dataset of NIR spectral data to skin lesions, as far as we know, a new dataset named NIR-SC-UFES, has been collected, annotated and analyzed generating the gold-standard for classification of NIR spectral data to skin cancer. The algorithms XGBoost, CatBoost, LightGBM and 1D-convolutional neural network (1D-CNN) were investigated to classify cancer and non-cancer skin lesions and they are also compared with the standard algorithms PLS-DA and SVM used in chemometrics.

The main contributions of this work are as in the following:
\begin{itemize}
\item We provide a new dataset named NIR-SC-UFES of the most common skin lesions not yet available in the literature.
\item We perform extensive experiments with the most powerful machine learning algorithms XGBoost, CatBoost, LightGBM and 1D-CNN to classify the NIR spectral dataset.
\item We investigate the influence of pre-processing and feature extraction as well data augmentation techniques to tackle small datasets.
\end{itemize}

The paper is organized as follows: Section 2 describes the data collection process of a new NIR spectral dataset for the most common skin lesions. In addition, we tackle the data pre-processing, feature extraction and data imbalance between classes. In Section 3, the machine and deep learning algorithms and the method to carry out the experiments are described. In Section 4, the results and discussions are presented. Section 5 ends up with conclusions and directions for future research.

%%% Início de seção. %%%
\section{Data and Preprocessing}

\subsection{The NIR-SC-UFES Dataset}
\label{descpro}
By means of a partnership with the Dermatological and Surgical Assistance Program (\textit{in Portuguese: Programa de Assistência Dermatológica e Cirúrgica}, PAD) at the University Hospital Cassiano Antônio Moraes (Hucam) of the Federal University of Espírito Santo (UFES), a dataset of skin lesions was created. The PAD-UFES-20 dataset \citep{Pacheco2020} is composed of the six most common skin diseases: 1) cancer, which consists of melanoma (MEL), basal cellular carcinoma (BCC) and squamous cellular carcinoma (SCC); and 2) non-cancer, which consists of nevus (NEV), actinic keratosis (ACK) and seborrheic keratosis (SEK).

\begin{figure}[ht]
        \centering
        \subfigure[BCC.]{
            \includegraphics[height=3cm, width=3cm]{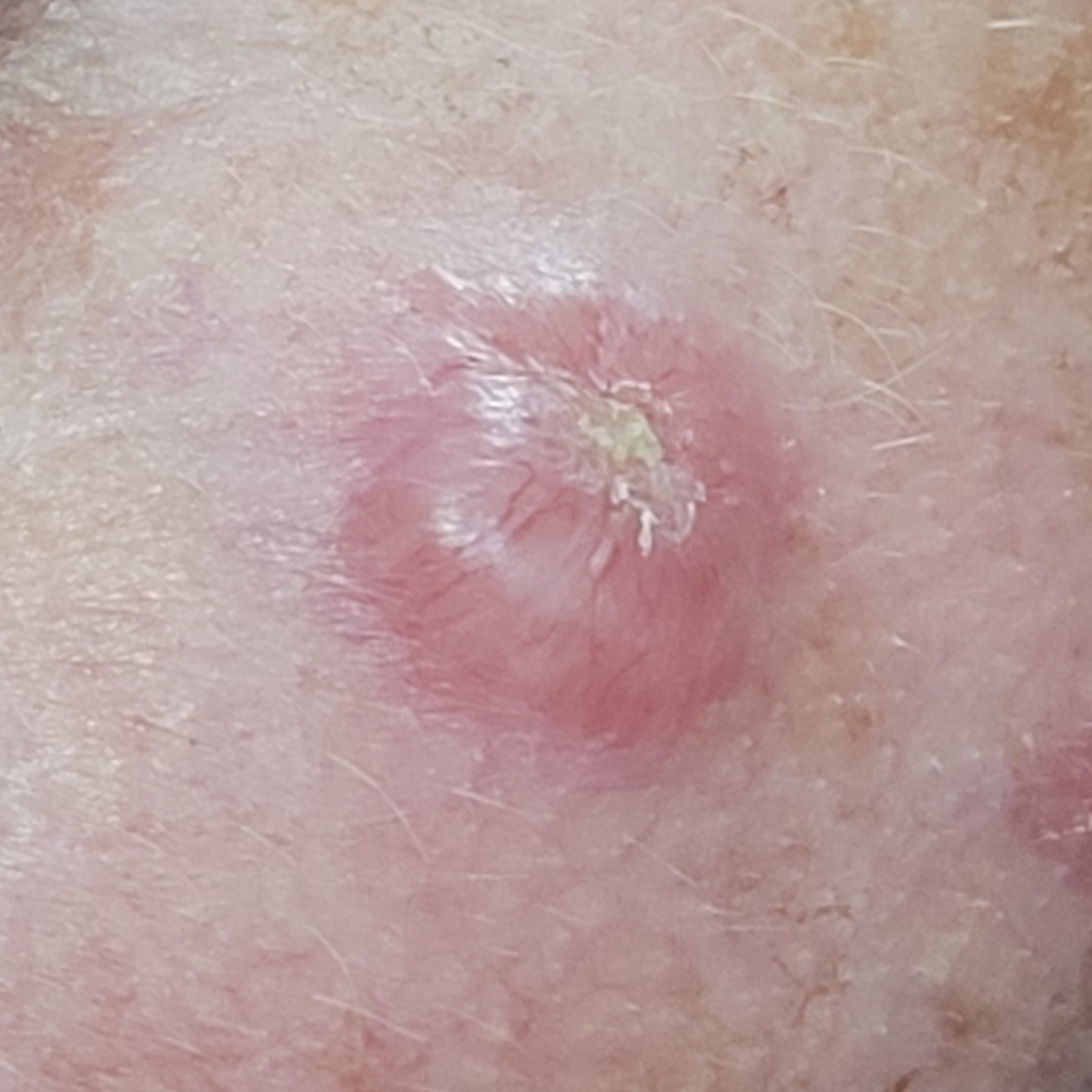}
        }
        \subfigure[SCC.]{
            \includegraphics[height=3cm, width=3cm]{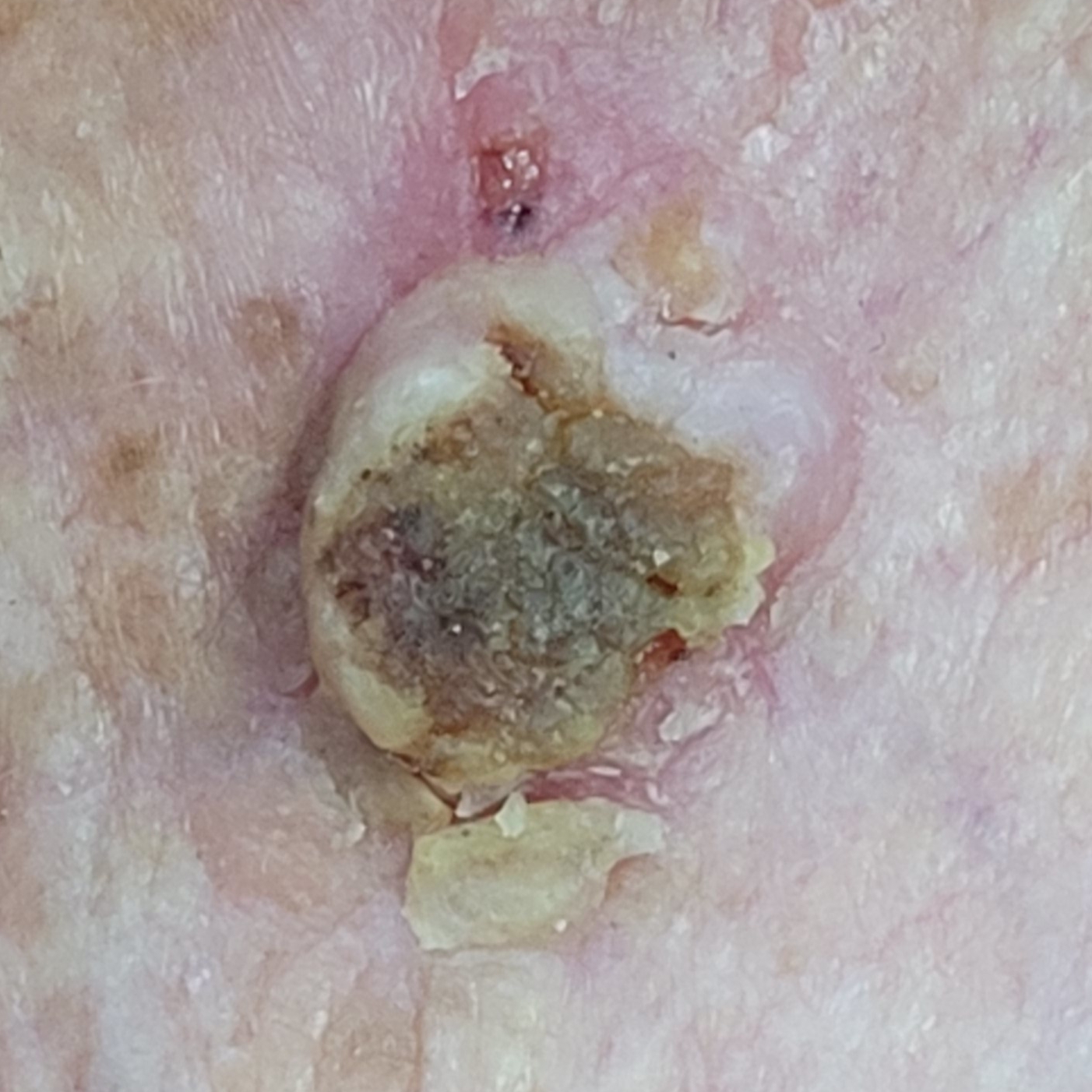}
        }
        \subfigure[ACK.]{
            \includegraphics[height=3cm, width=3cm]{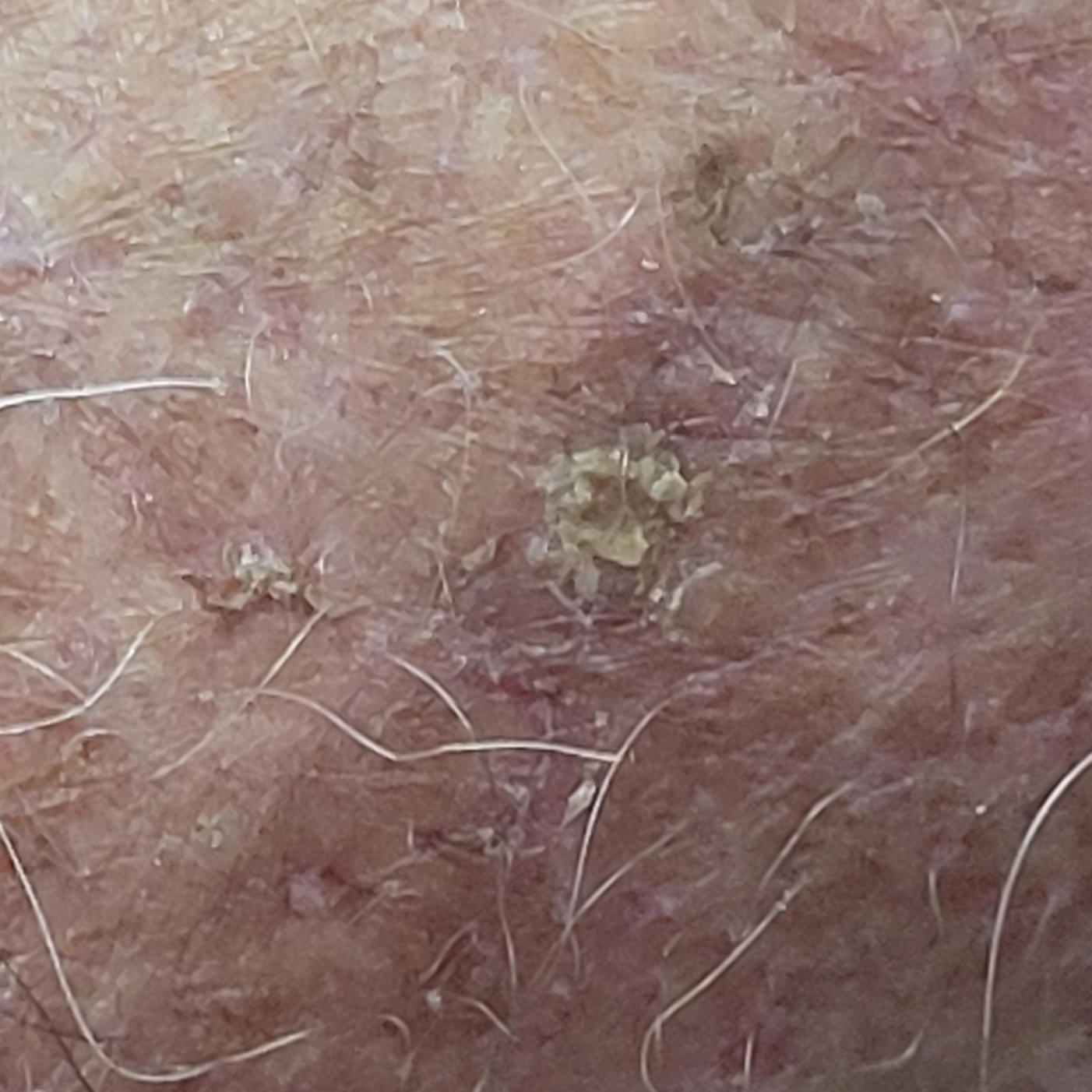}
        }

        \subfigure[MEL.]{
            \includegraphics[height=3cm, width=3cm]{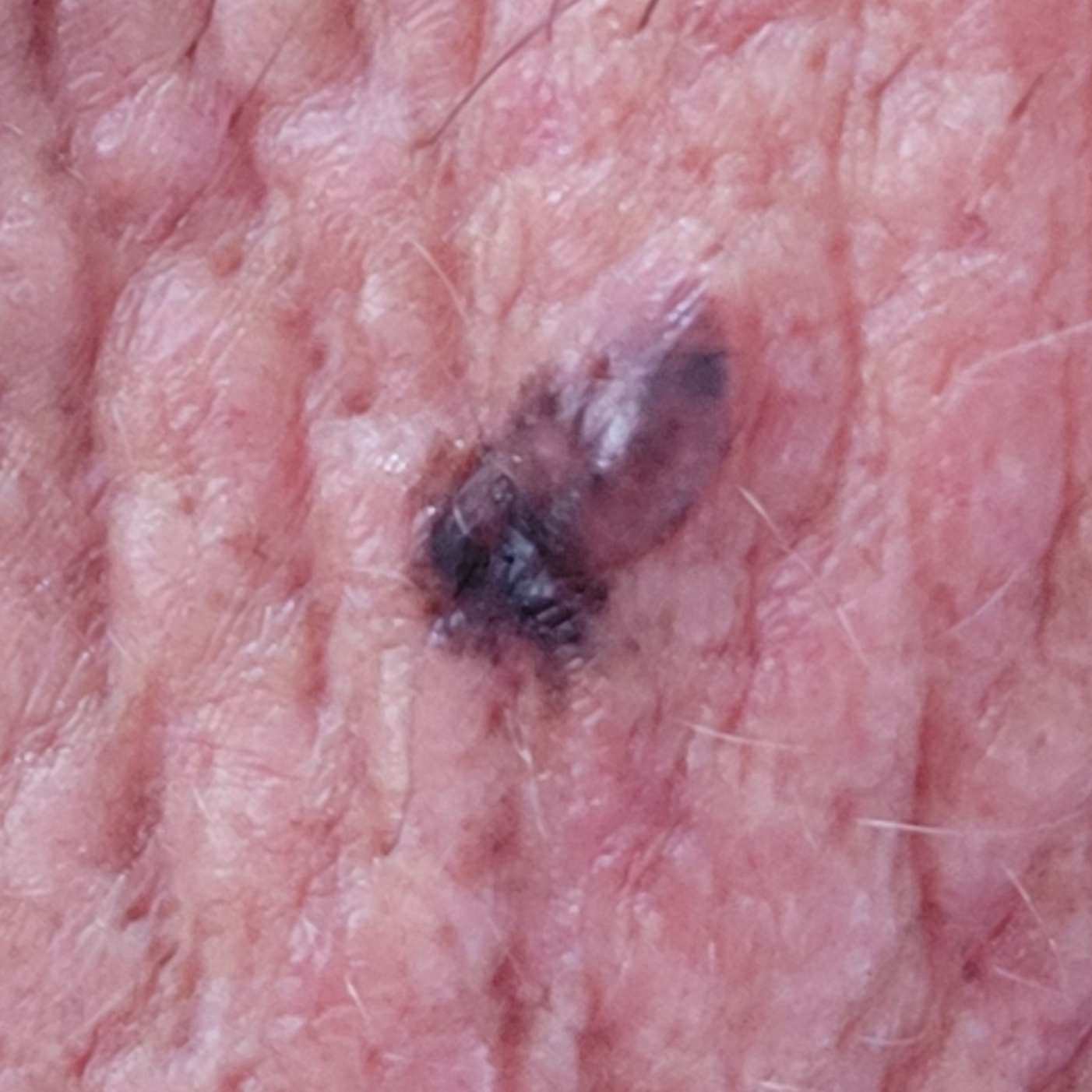}
        }
        \subfigure[NEV.]{
            \includegraphics[height=3cm, width=3cm]{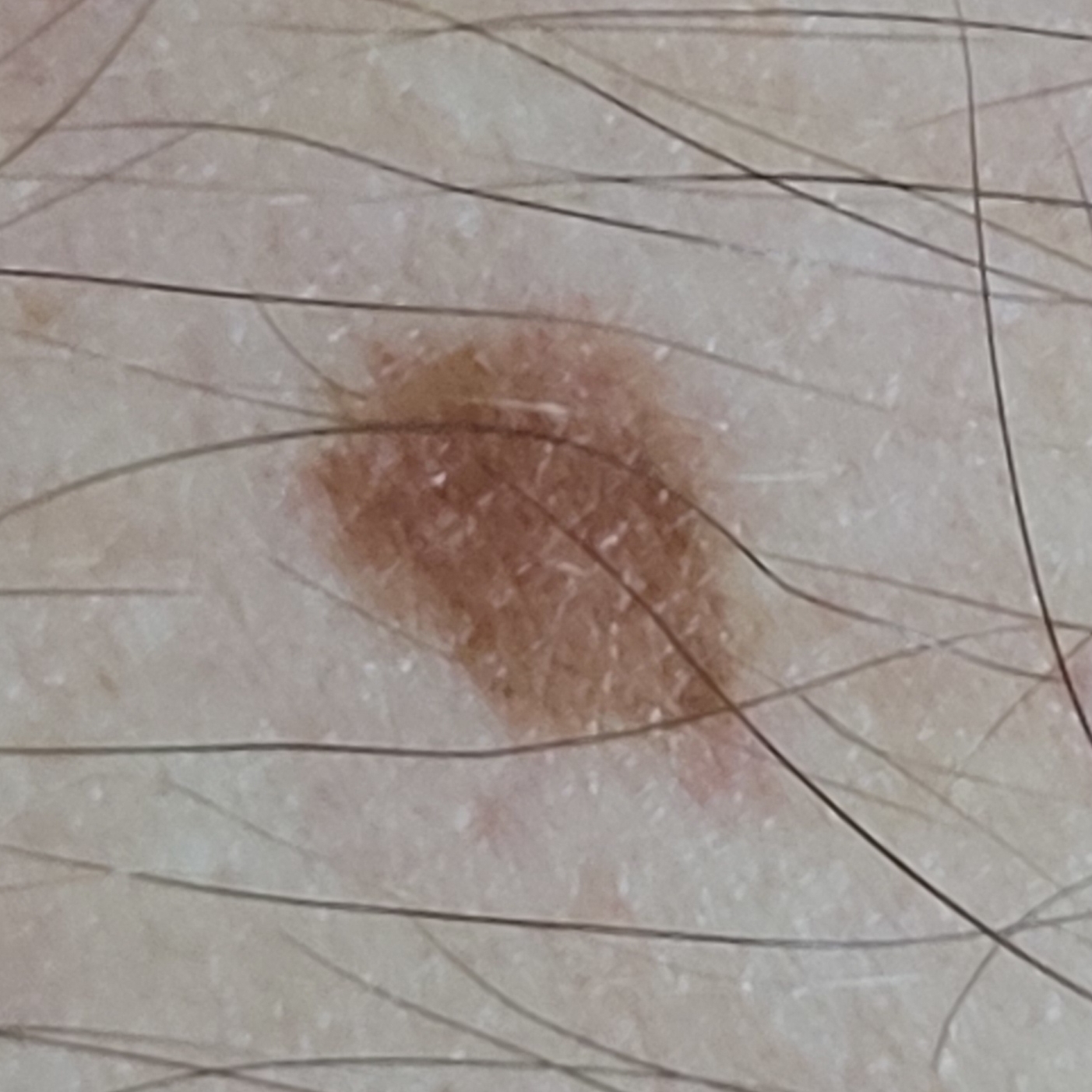}
        }
        \subfigure[SEK.]{
            \includegraphics[height=3cm, width=3cm]{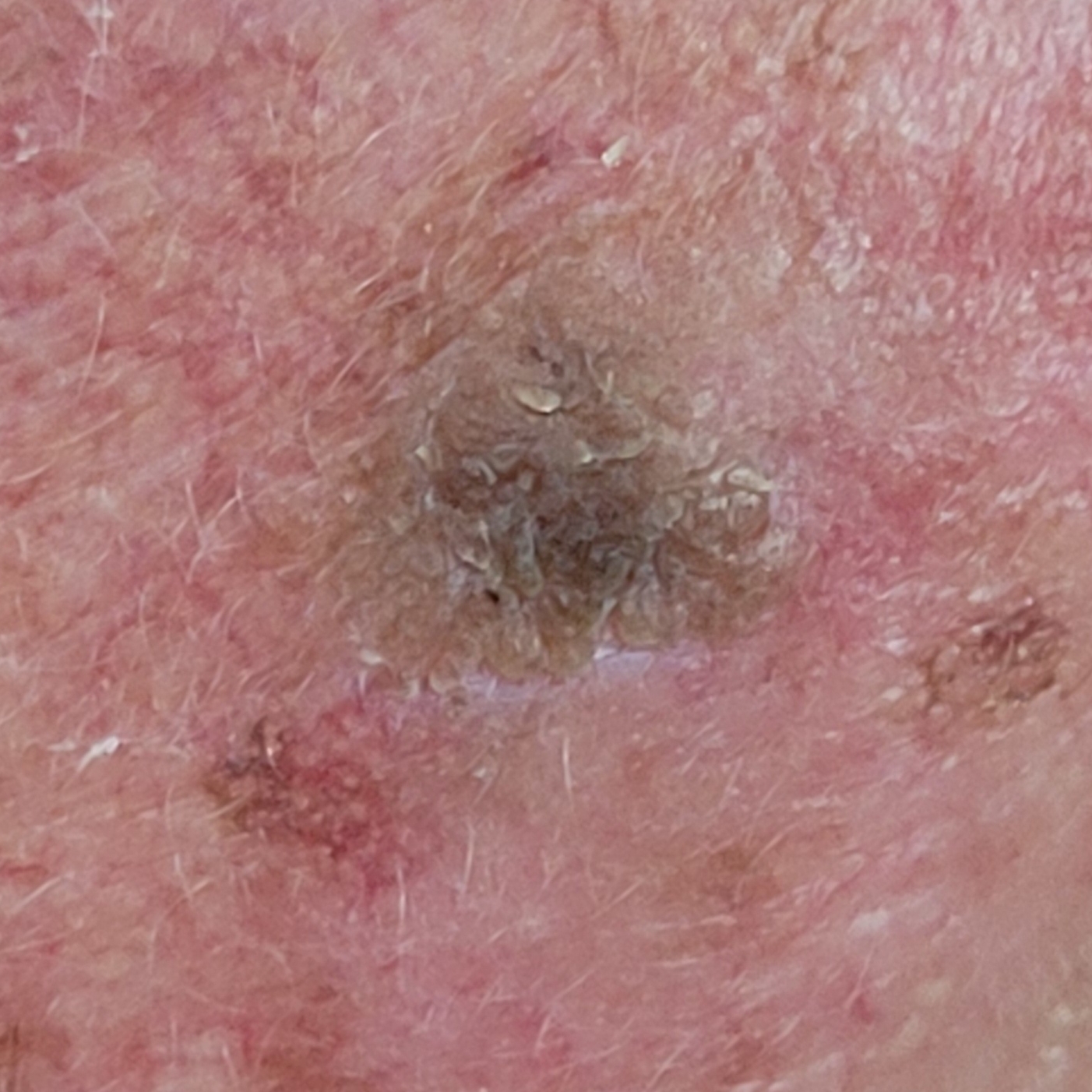}
        }
        \caption{A photo sample of each type of skin lesion investigated in this work.}
        \label{photo_lesions}
    \end{figure}

 The PAD-UFES-20 dataset consists of images taken from smartphones and metadata (patient and collected lesion information). Despite the importance of images in automated diagnostics, there are specific lesions such as MEL and NEV that are harder or even impossible to detect cancerous patterns using clinical images or patient lesion information \citep{Pacheco2019}.  Studies \citep{Pacheco2019} were carried out to automatically recognize the main skin lesions. However, there are still difficulties in correctly discriminating all types of skin lesions, especially in differentiating between MEL and NEV and between MEL and both the keratosis (SEK and ACK).  Figure \ref{photo_lesions} shows the most common six skin lesions investigated in this work.

\citet{McIntosh2002} investigated the use of Near-Infrared (NIR) spectroscopy by differentiating some types of skin lesions such as NEV and MEL. Spectroscopy measures the amount of light reflected at such wavelengths, when incident on the molecule to be analyzed, being a non-invasive and fast approach \citep{stuart2005}. 

\begin{figure}[ht]
    \includegraphics[width=0.3\linewidth]{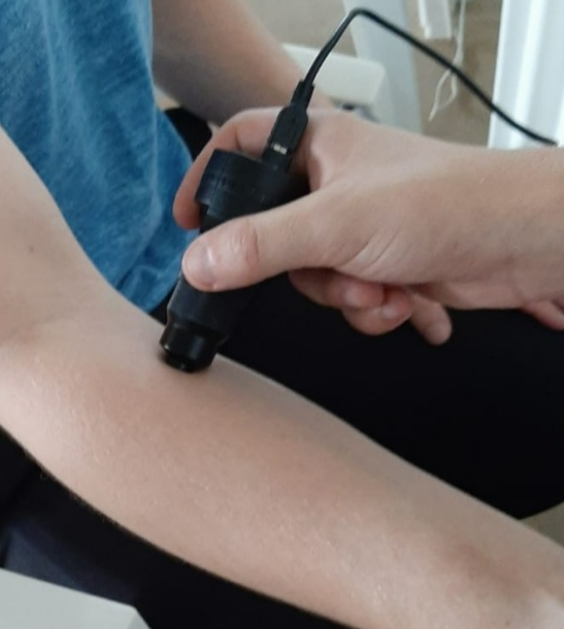}
    \centering
    \caption{Acquisition of the NIR spectral data of a patient lesion using the Micronir portable spectrometer.}
    \label{coleta_nir}
\end{figure}

In the search for an alternative source of information to detecting skin cancer, a new dataset, named NIR-SC-UFES, was collected using near infrared spectroscopy alongside the PAD program. From March, $2021$ to December, $2022$, the collection of the spectral data of skin lesions was carried out using the Micronir spectrometer. The number of dimensions of each sample consists of $125$ amplitude values in the wavelength range of $900$ nm to $1700$ nm, each one representing the average measured in an interval of $6.4$ nm. Figure \ref{coleta_nir} shows the process of data collection using the portable Micronir spectrometer.

\FloatBarrier

For benign skin lesions the diagnostic is performed by a committee of 3 dermatologists and for malignant skin lesions the result of the biopsy is the gold-standard. For the classification of skin lesions, two different labels will be considered, the non-cancer, which contains spectra of the samples of ACK, SEK and NEV; and cancer referring to the spectra of samples of BCC, SCC and MEL. A well known problem in the skin cancer diagnosis using CAD is the low number of samples and the imbalance between classes. The NIR-SC-UFES dataset contains $586$ samples of benign lesions and $385$ of malignant, totaling $971$ samples. This corresponds to 60.4\% of the samples are non-cancer and 39.6\% are cancer. Counting each lesion specifically, the dataset contains 336 ACK samples, 188 SEK, 62 NEV, 302 BCC, 72 SCC and 11 MEL.

\begin{figure}[ht]
    \includegraphics[width=0.9\linewidth]{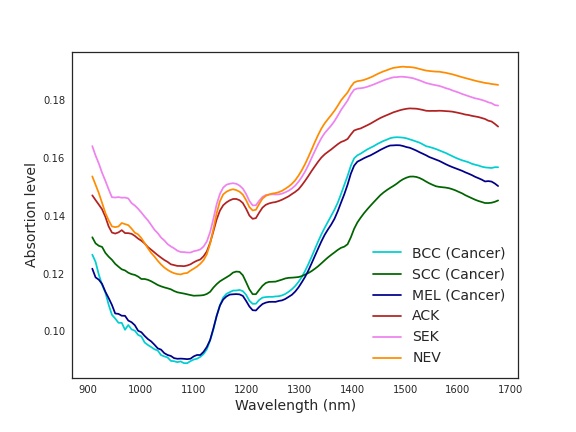}
    \caption{Spectral data sample of the six kind of skin lesions contained in the NIR-SC-UFES dataset.}
    \label{spectrum_multiclass}
\end{figure}

Figure \ref{spectrum_multiclass} shows a spectral data sample of the six kind of skin lesions contained in the NIR-SC-UFES dataset. 

%%%%%%%%%%%%%%%%%%%%%%%%%%%%%%%%
%%%%%%%%%%%%%%%%%%%%%%%%%%%%%%%%
%%%%%%%%%%%%%%%%%%%%%%%%%%%%%%%%

In the field of chemometrics, it is commonly used traditional algorithms to work with spectral data, such as \textit{Partial Least Squares} (PLS) \citep{BRERETON201590}. However, the use of machine learning and deep learning algorithms are increasingly used for this type of task, providing superior results to those obtained by standard algorithms of chemometrics \citep{Yuanyuan2018}. Some of these models may face greater difficulty in generalizing with a reduced amount of training data, due to their architecture and training method. In this paper, we apply the most powerful machine learning algorithms XGBoost, CatBoost, LightGBM and 1D convolutional neural networks to the NIR-SC-UFES dataset to classify skin cancer. In addition, we investigate the effects of pre-processing and data augmentation methods in the performance of the algorithms.

\subsection{Pre-processing}
Data coming from a spectrometer are susceptible to various types of noise, whether due to external interference or the precision of the equipment \citep{Smulko2015}. So, data pre-processing techniques are applied to the spectral data. The dispersion of the radiation is dependent on the physical nature of the sample particles. For the correction of such divergences, techniques of centralization of each individual sample are used, being the \textit{Standard Normal Variate} (SNV) highly used for the removal of such components \citep{snv}.

The SNV value is calculataded by:

\[SNV = (y_i - \bar{y})/\sqrt{\sum_{i=1}^{n}{{(y_i - \bar{y})}^{2}}/{(n - 1)}} \]
where $n$ is the number of wavelengths in the sample, $y_i$ is the value of the absorbance at each wavelength $i$ for each sample and $\bar{y}$ is the average of the amplitudes of the wavelengths.

\subsection{Feature Extraction}
\label{excarac}
\citet{Arajo2021} presented a feature extraction method for Raman spectral data of skin lesions that provide superior results compared to experiments without feature extraction. The method consists of calculating statistical features in a subsequence of the original signal of the sample. Some parameters need to be defined, such as the subsequence size and the percentage of overlapping between subsequences. The subsequence generated features that were used in this work are listed in Table \ref{features}, where $S_{i,m}$ represents a subsequence of the signal, with $ i$ the starting point of the substring and $m$ its length.

In order to find the optimal number of subsequences, we optimize the hyper-parameters using TPE (Tree-structured Parzen Estimator) \citep{Bergstra2011} as sampler. The optimization algorithm was used along with the framework Optuna \citep{Akiba2019}, optimizing for $200$ epochs each validation set. The goal is to automatically find out the number of windows and what features generate the best results (for instance, the highest accuracy or recall). For the number of windows, a minimal value of 5 and maximum of 50 was pre-specified.
\clearpage

\begin{table}[ht]
\vspace{-8.5em}%
\centering
\begin{tabular}{m{90pt}|m{100pt}|m{155pt}}
\hline
\textbf{Feature} & \textbf{Description}                                                               & \textbf{Formula} \\
\hline
Mean    & Simple average of the subsequence.                                          &   $\mu_{i,m} = \frac{1}{m}\sum_{k=0}^{m-1}s_{i+k_r}$ \\
\hline
Median             & Value that separates the major and minor half of a sample.          & $M_{m} = S_{\frac{N+1}{2}}$ \\ 
\hline
Standard Deviation       & Measures the amount of spreading of the subsequence.                      &   $\sigma_{i,m} = \sqrt{\frac{1}{m}\sum_{k=0}^{m-1}s_{i+k_r} - \mu_{i,m} }$      \\
\hline
Kurtosis            & Measures the amount of spreading of the subsequence. &  $ku_{i,m} = \frac{\frac{1}{m}\sum_{k=0}^{m-1}(s_{i + k_r} - \mu_{i,m})^4}{(\sigma_{i,m})^4}$       \\
\hline
Skewness          & Measures the lack of symmetry in the distribution of subsequence values.   &   $sk_{i,m} = \frac{\frac{1}{m}\sum_{k=0}^{m-1}(s_{i + k_r} - \mu_{i,m})^3}{(\sigma_{i,m})^3}$      \\
\hline
Maximum              & Maximum value of the subsequence.                                           &   $Max_{i,m} = max(S_{i,m})$      \\ 
\hline
Minimum              & Minimum value of the subsequence.                                           &  $Min_{i,m} = min(S_{i,m})$ \\      
\hline
Peak                & Measures the peak value of a subsequence.                    & $P_{m} = max(|S_{i}|)$ \\
\hline
Peak to Peak        & Measures the distance from a signal peak to the valley.                      & $P_{k} = max(S_{i}) - min(S_{i})$ \\
\hline
RMS(Root Mean Square)        & Measures the square root of the mean square of the set.                    & $RMS = \sqrt{\frac{1}{m}\sum_{i=1}^{m}S_{i}^{2}}$ \\
\hline
Variance           & Measurement of statistical dispersion of the signal.                              & $Var = \frac{\sum_{i=1}^{m}(S_{i}-\mu)^{2}}{m-1}$ \\
\hline
Crest Factor     & Shows the ratio between the peak values and the actual waveform value.     & $CF = \frac{P_{m}}{RMS}$ \\
\hline

\end{tabular}
\caption{Formulae to calculate the extracted features for each subsequence $S_{i,m}$ \citep{Arajo2021}.}
\label{features}
\vspace{-58pt}
\end{table}
\clearpage

\subsection{Oversampling with SMOTE}
Imbalanced data can generate problems such as biased validation metrics, and models with greater preference for a class, since samples of the majority class was more present in the training data. In order to tackle this issue we use oversampling, which aims to generate synthetic samples of the minority class, in order to balance the data. It is worth mentioning that creation of synthetic data is a common practice in spectroscopy \citep{alli2020}.

The Synthetic Minority Over-sampling Technique (SMOTE), introduced by \citet{smote}, is a method used for balancing imbalanced datasets. In this approach, the first step is to calculate the number of observations that need to be generated to balance the data. Typically, this number is determined in a way that equalizes the size of the minority class with that of the majority class. By applying SMOTE, synthetic samples are created based on the characteristics of the minority class to increase its representation in the dataset. This helps address the class imbalance issue and improves the performance of machine learning models on such datasets. Next, each iteration selects a random observation from the minority class and with it, the K-Nearest Neighbors is applied, where for each of these neighbors a distance metric is calculated. Finally, the difference between the distances of the instances is obtained, and multiplied by a random number in the interval [0, 1], thus obtaining the position in the feature vector of the generated instance. Algorithm \ref{alg1} lists the pseudo-code for synthetic sample generation using SMOTE.

% \begin{algorithm}[Ht]
%     \caption{Pseudocode for the SMOTE algorithm}
%     \label{alg1}
%     \SetAlgoLined
%     \KwData{ \\
%     Number of minority class samples $T$; \\
%     Amount of SMOTE \textit{N\%}  \\
%     Number of nearest neighbours \textit{k} \\
% } 
%     \textit{numattrs} = Number of atributes \\
%     \textit{Sample}: Original vector of minority class \\
%     \textit{newindex}: Synthetic sample counter, starting from 0 \\
%     \textit{Synthetic}: Synthetic data vector \\
%     \For{\textit{i} = 1 \textit{to} \textit{T}}{
%         Calculate the \textit{k} nearest neighbours to \textit{i}, and save the indexes in \textit{nnarray} \\
%         \While{$N \neq 0$}{
%             Choose a number between 1 and \textit{k}, call it \textit{nn}. This step chooses \textit{k} nearest neighbours \textit{i}. \\
%             \For{\tewxtit{attr} = 1 \textit{to} \textit{numattrs}}{
%                 \textit{dif} = \textit{Sample}[\textit{nnarray}[\textit{nn}]][\textit{attr}] − \textit{Sample}[\textit{i}][\textit{attr}] \\
%                 \textit{gap} = random number between 0 and 1 \\
%                 \textit{Synthetic}[\textit{newindex}][\textit{attr}] = \textit{Sample}[\textit{i}][\textit{attr}] + \textit{gap} \ast \textit{dif} \\
%                 }
%             \textit{newindex}++ \\
%             \textit{N} = \textit{N} - 1   \\
%             }
%     }
% \end{algorithm}

\begin{algorithm}[ht]
    \caption{Pseudocode for the SMOTE algorithm}
    \label{alg1}
    \begin{algorithmic} [5]       
        \Require {Number of minority class samples $T$;  
         Amount of SMOTE \textit{N\%};
         Number of nearest neighbours \textit{k}.}

        \State \textit{numattrs} = Number of atributes
        \State  \textit{Sample}: Original vector of minority class
        \State  \textit{newindex}: Synthetic sample counter, starting from 0
        \State \textit{Synthetic}: Synthetic data vector
        \For{\textit{i} = 1 \textit{to} \textit{T}}
            \State $ \textrm{Calculate the \textit{k} nearest neighbours to \textit{i}.}$
            \State $ \textrm{Save the indexes in \textit{nnarray}.}$
            \While {$N \neq 0$}
                \State $\textrm{Choose a number between 1 and \textit{k}, call it \textit{nn}.}$
                \State $\textrm{This step chooses \textit{k} nearest neighbours \textit{i}.} $
                \For{$attr = 1 \to numattrs$}
                    \State $dif = Sample[nnarray[nn]][attr] - Sample[i][attr] $
                    \State $gap = \textrm{random number between 0 and 1}$ 
                    \State \textit{Synthetic}[\textit{newindex}][\textit{attr}] = \textit{Sample}[\textit{i}][\textit{attr}] + \textit{gap} * \textit{dif} 
                \EndFor
                \State $newindex++$
                \State $N = N - 1$
            \EndWhile
        \EndFor
    \end{algorithmic}
\end{algorithm}

\subsection{GAN-based Data Augmentation}

 Generative Adversarial Networks (GANs) is a promising data augmentation technique introduced by \cite{goodfellow2014generative}. GANs consist of two networks that compete with each other, the generator network $G$ and the discriminator network $D$. The objective of $G$ is to learn how to generate synthetic data $G(Z)$ similar to the real ones from noise $Z$, thus the synthetic and real data are used as input in $D$, being responsible for performing the judgment of the data received, informing whether they are true or false. This competition is described by the loss function:

\[\min_{G} \max_{D}\mathcal{L}_{GAN}(G,D) = E_{x}[\log D (x)] + E_{Z}[\log (1 - D (G(Z))) \] which causes the GANs to generate new samples that differ from the original set. 

\begin{figure}[ht]
    \includegraphics[width=0.9\linewidth]{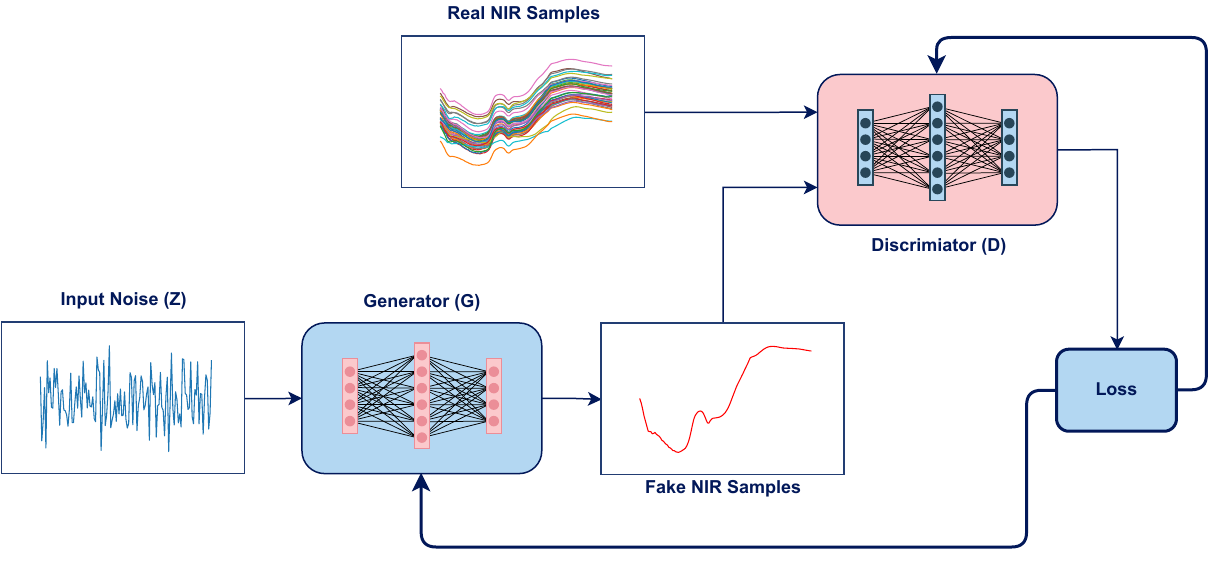}
    \caption{Structure of the GAN network for generating NIR samples. A noise vector $Z$ passes through the generator ($G$) obtaining the generated output, and the discriminator $D$ is simultaneously trained to distinguish the generated signals from the real ones. The reconstruction loss measures how close the generated signals are to the real ones.}
    \label{gan}
\end{figure}

\cite{PAVLOU2022104634} proposed a simple GAN scheme to generate synthetic Raman spectra using data from both healthy and osteoporosis bone samples \textit{ex vivo}. Their scheme works as in the following: the generated data by the GAN and real data are normalized and applied Principal Component Analysis (PCA), where ellipses are generated with a certain confidence interval to evaluate the variability of the augmented data set and data that are outside of the ellipse are considered outliers and consequently discarded. Since this scheme is of general purpose it can applied to augment any one-dimensional biomedical signal. So, this data augmentation technique was also used in our experiments for generating NIR spectral data for skin lesions aiming to increase the number of samples from the minority class (cancer) to the same size as the majority class (non-cancer). The process works in the following way: the raw data of the NIR spectra referring only to the cancer class were used to train the GAN and obtain new synthetic malignant spectra from the real ones. Next, the real and generated data were normalized with the SNV technique to perform the variability analysis and discard possible outliers. The confidence interval used for the ellipses was $95\%$, i.e., the ellipse that is drawn around the two axes (major and minor) is supposed to contain the $95\%$ of the data points. The GAN structure for our case is shown in Figure \ref{gan}.

\section{Algorithms and Methods}
Next, we shortly describe the machine learning algorithms XGBoost, CatBoost, LightGBM,  1D-CNN  as well as PLS-DA and SVM used in this work.

\subsection{Algorithms}

\subsubsection{Extreme Gradient Boosting (XGBoost)}

The XGBoost algorithm \citep{Chen2016} consists of a cast of simple predictors (similar regression trees) also called a weak learner, in which the responses of each individual predictor are aggregated. The set of regression trees that make up the cast is elaborated in two steps: (a) a tree $T_{0}$ is built by adding to its structure the branch of the attribute that most improves the predictor at each step and, once its structure is defined, the optimal values are calculated for the $l$ leaves of the tree and (b) the inclusion of new trees to the set is guided by the residual error of the model, so that new predictors complement the imperfections of the previous ones \citep{Friedman2002}. Unlike other boosting techniques, Extreme Gradient Boosting is a model that makes use of regularization and overfitting control to improve the performance of the algorithm.

\subsubsection{Categorical Boosting (CatBoost)}

The CatBoost algorithm  \citep{Prokhorenkova2018} was developed with the purpose of easily handling large datasets as well as heterogeneous data. CatBoost is a combination of the names \say{category} and \say{boosting} as it handles categories on its own and is based on the gradient boosting algorithm. CatBoost only requires indexes of categorical features to perform the transformation of categories into numerical data \citep{Hancock2020}, unlike the vast majority of data mining algorithms that require several pre-processing steps. Furthermore, another relevant feature of CatBoost is that it does not require large datasets for training.

\subsubsection{Light Gradient Boosting Machine (LightGBM)}

The LightGBM algorithm \citep{Ke2017} was developed to overcome the limitations of the Gradient Boosting Machine (GBM) algorithm in big data problems because the GBM becomes very inefficient due to its construction being carried out according to the number of instances and attributes. Thus, a sampling technique is necessary, i.e., operating with smaller samples to reduce complexity. The LightGBM algorithm implements the sampling technique by the Gradient-Based One-Side Sampling (GOSS) method, since it was observed that different instances of the data present different gradients, and these in turn act in different ways to gain information. Those with greater gradient in absolute values result in greater information gain.

The algorithm seeks to carry out the sampling in order to obtain only instances with the highest gradient, allowing a significant reduction in the data while maintaining a significant part of the information. Another advantage of this sampling is that it allows to reduce noise by keeping only data with high gradients. As a result, the LightGBM algorithm performs faster than traditional GBM algorithms and achieves accuracy equivalent to or greater than those counterparties.

\subsubsection{Artificial Neural Networks}

A feedforward neural network consists of information processing elements called artificial neurons, which are arranged in layers and are interconnected with each other \citep{nn}. Neural networks are characterized by three types of layers: the input layer, whose neurons receive the primary data directly, the hidden layers that contain the intermediate neurons and perform most of the interconnections between them, and the output layer where, after all the information flows through the network, it assumes the role of the final transformation of the data and represents the output of the network.

Convolutional neural networks differ from feedforward neural networks by having convolution and pooling layers in at least one of its layers, being able to extract spatial and temporal characteristics from data, such as images and time series \citep{goodfellow}. The convolutional layers of one dimension are used due to the unidimensional characteristic of the spectra, and pooling layers to reduce the dimensionality of the output of the convolutional layer \citep{goodfellow}. After feature extraction with the convolution kernels and dimensionality reduction, feedforward networks are used with ReLU activation functions \citep{relu} and neurons at the output layer with sigmoid activation function. To generate a CNN architecture, convolution and pooling layers were used, followed by a feedforward layer for mapping the relationships between the variables.

\subsubsection{Standard Algorithms Used for Spectral Data}

Partial Least Squares-Discriminant Analysis (PLS-DA) is a supervised classification method developed based on regression modeling. It is a multivariate dimensionality  reduction technique that has been standard in the field of chemometrics over two decades \citep{Gottfries1995}.

Support Vector Machines (SVM) is a supervised learning method and performs well in linear and non-linear modeling due to the kernel mapping that transforms the sample space \citep{svm}. In this study, different types of kernel functions were tested, such as radial basis function, linear and polynomial. For better performance of the SVM, the TPE \citep{Bergstra2011} was used to optimize three parameters, the cost constant, $C$, which optimizes the slack variable, the tolerance, $\epsilon$, which delimits the slack, and the gamma constant, $\gamma$, which optimizes the kernel function.

\subsection{Experimental Methodology}

In the following, we describe the experimental methodology to carry out the experiments.

\subsubsection{Cross-Validation}

Data separation was carried out as follows: from the 971 samples, 776 samples for training (80\%) and 195 for testing (20\%). This represents a percentage of $60.35\%$, and $39.65\%$ for non-cancer and cancer samples,  respectively. By counting each lesion specifically, the dataset contains 336 ACK, 188 SEK, 62 NEV, 302 BCC, 72 SCC, and 11 MEL samples. The data were stratified according to the amount of injury, ensuring that we have, e.g., 9 MEL ($80\%$) in training and 2 MEL ($20\%$) in testing.

\begin{figure}[ht]
    \includegraphics[width=0.9\linewidth]{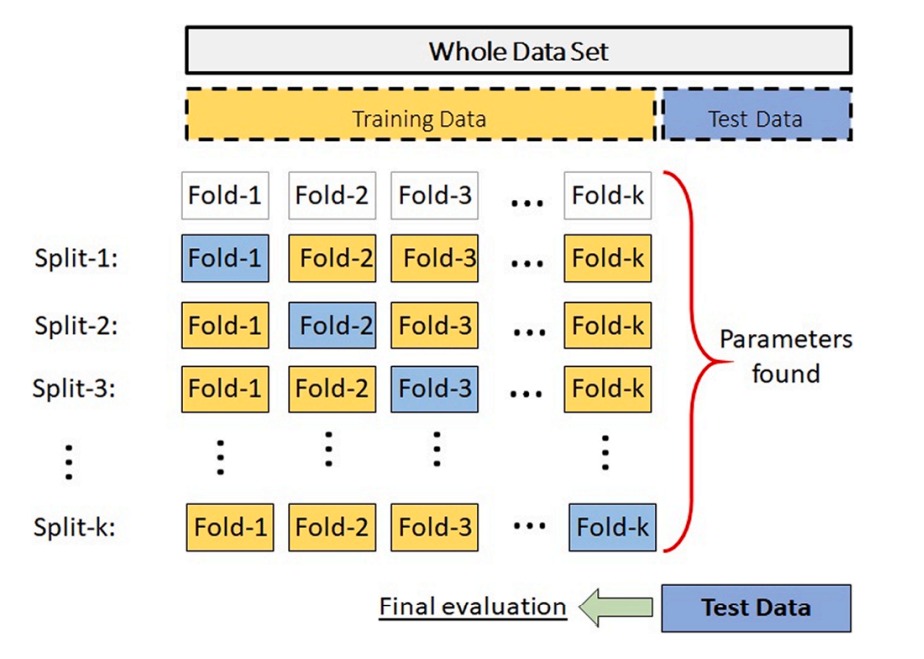}
    \caption{Cross Validation performed using k-fold splits.}
    \label{kfold}
\end{figure}

To separate the training data, the Cross-Validation (CV) technique was used. CV is used to avoid methodological errors, it separates the data into $k$ equal parts named folds, with $(k-1)$ parts used as training data and $1$ for validation. This separation is repeated $k$ times, with the validation part at each partition a shift is performed, so that in the end, the entire data set was used in the process. In our experiments, $k = 5$ was used, a default value present in the \textit{sci-kit} library. The CV separates the data between $80\%$ training samples and $20\%$ validation samples into each of the $k$ parts, maintaining stratification based on the skin lesions. Both the separation between training and testing, and that of the CV for training and validation, a seed was used to keep the choice of folders by default for each test. Figure \ref{kfold} shows a diagram of the dataset split into train and test, and the train dataset split in $k$ folds.

\subsubsection{Metrics}

To evaluate the performance of the classification algorithms, five common metrics in the literature were used. Accuracy, Balanced Accuracy, Recall, Precision, and F-score respectively, are calculated by:

\begin{equation}
    ACC = \frac{TP + TN}{TP + FP + TN + FN}
    \label{ACC}
\end{equation}

\begin{equation}
    BACC = \frac{\frac{TP}{TP + FN} + \frac{TN}{TN + FP}}{2}
    \label{BACC}
\end{equation}

\begin{equation}
    Recall = \frac{TP}{TP + FN}
    \label{RE}
\end{equation}

\begin{equation}
    Precision = \frac{TN}{TN + FP}
    \label{PR}
\end{equation}

\begin{equation}
    \textit{F-score} = 2* \frac{Recall * Precision}{Recall + Precision}
    \label{FS}
\end{equation}

The variables \textit{TP, TN, FP}, and \textit{FN} are abbreviations of True Positive, True Negative, False Positive, and False Negative, respectively. 
For skin lesions classification, it is preferable that the classifier correctly matches the class of cancer lesions, as the patient's condition may be critical and the search for treatment must be immediate.

\subsubsection{Hyperparameters setting}

Among the algorithms used, XGBoost, CatBoost, LightGBM and 1D-CNN have hyperparameters, which are adjusted to obtain the best performance of the models. The search for optimal hyperparameters was performed using the training set. Next, the best hyperparameter combination was used to train and evaluate the model.

Table \ref{hiperparametrobusca} presents the hyperparameter search space for XGBoost, CatBoost and LightGBM algorithms.

\begin{table}[!htbp]
\centering
\begin{tabular}{l|m{150pt}|l}
\hline
Algorithm & Hyperparameter      & Search space               \\ \hline
XGBoost   & class weight       & {[}1 $\sim$ 25{]}    \\
XGBoost   & max depth         & {[}1 $\sim$ 15{]}               \\
XGBoost   & number of trees       & {[}10 $\sim$ 100{]}              \\
XGBoost   & learning rate  & {[}0.01 $\sim$ 1{]}                 \\
XGBoost   & col sample by tree & {[}0.5 $\sim$ 1{]}  \\
XGBoost   & subsample & {[}0.1 $\sim$ 1{]}  \\
XGBoost   & alpha regularization & {[}0 $\sim$ 20{]}  \\
XGBoost   & lambda regularization & {[}0 $\sim$ 20{]}  \\
\hline
CatBoost   & max depth          & {[}1 $\sim$ 16{]}   \\
CatBoost   & l2 leaf regularization         & {[}1 $\sim$ 40{]}   \\
CatBoost   & number of trees         & {[}1 $\sim$ 200{]}   \\
CatBoost   & colsample by level         & {[}0.1 $\sim$ 1{]}   \\
CatBoost   & learning rate         & {[}0.01 $\sim$ 1{]}   \\
\hline
LightGBM   & class weight       & {[}1 $\sim$ 15{]}   \\
LightGBM   & number of trees       & {[}10 $\sim$ 100{]}   \\
LightGBM   & learning rate  & {[}0.01 $\sim$ 1{]}   \\
LightGBM   & leaf number  & {[}20 $\sim$ 5000, step=20{]}    \\
LightGBM   & max depth          & {[}3 $\sim$ 12{]}   \\

\end{tabular}
\caption{Hyperparameter search space for XGBoost, CatBoost and LightGBM algorithms.}
\label{hiperparametrobusca}
\end{table}

For the problem at hand, the 1D-CNN architecture used,  consists of up to three layers of convolution and pooling, followed by a fully connected layer. The search space of the hyperparameters of the 1D-CNN network is presented in Table \ref{hiperparametrocnn}.

\begin{table}[htbp]
\centering
\begin{tabular}{l|m{200pt}|l}
\hline
Algorithm & Hyperparameter      & Search space               \\ \hline
1D-CNN       & learning rate             & {[}0.01, 0.001, 0.0001{]} \\
1D-CNN       & batch size               & {[}16 $\sim$ 128{]}     \\
1D-CNN       & number of convolutional layers                   & {[}1, 2, 3{]}              \\
1D-CNN       & numbers of neurons on convolutional layers          & {[}64 $\sim$ 256{]}               \\
1D-CNN   & numbers of neurons on deep layers           & {[}64 $\sim$ 256{]}               \\
\end{tabular}
\caption{Hyperparameter search space for the 1D-CNN.}
\label{hiperparametrocnn}
\end{table}

\section{Experimental Results}
In this work  we investigate the performance of the machine learning algorithms XGBoost, CatBoost, LightGBM, 1D-CNN as well as the standard algorithms SVM and PLS-DA applied to the new NIR-SC-UFES dataset. in order to address the problem of imbalanced data, two oversampling techniques SMOTE and GAN are used. The experiments are carried out for the 3 sets of data:

\begin{enumerate}[label=\Roman*]
   \item- Raw spectral data,
    \item- Balanced data using SMOTE,
    \item- Balanced data using GAN.
\end{enumerate}

For each dataset the experiments are carried out as follows:

\begin{enumerate}[label=\alph*]
   \item- Using original spectral data without pre-processing,
    \item- Using spectral data pre-processed with SNV,
    \item- Using data pre-processed with SNV  and with feature extraction.
\end{enumerate}

The results obtained are presented in the following.

\subsection{Results obtained using raw data}

Table~\ref{exp-1} presents the results of the experiments without pre-processing. In bold is highlighted
the highest average for each metric and standard deviation.

\begin{table}[htbp]
\centering
\resizebox{\textwidth}{!}{\begin{tabular}{l|l|l|l|l|l}
\hline
Algorithm & ACC               & BACC           & Recall          & Precision      & F-Score \\ \hline
PLS-DA    &  0.743 $\pm$ 0.004   &  0.699 $\pm$ 0.005      &  0.494 $\pm$ 0.011 & 0.773 $\pm$ 0.005 & 0.602 $\pm$ 0.009    \\
SVM       &  \textbf{0.833 $\pm$ 0.013}   &  \textbf{0.809 $\pm$ 0.015}      &  0.693 $\pm$ 0.033 & \textbf{0.857 $\pm$ 0.027} & \textbf{0.766 $\pm$ 0.021}  \\
XGBoost   &  0.740 $\pm$ 0.017   &  0.741 $\pm$ 0.013      &  \textbf{0.748 $\pm$ 0.027} & 0.649 $\pm$ 0.029 & 0.694 $\pm$ 0.012 \\
CatBoost  &  0.771 $\pm$ 0.021   &  0.756 $\pm$ 0.023      &  0.683 $\pm$ 0.045 & 0.724 $\pm$ 0.031 & 0.702 $\pm$ 0.029 \\
LightGBM  &  0.760 $\pm$ 0.030   &  0.757 $\pm$ 0.025      &  0.743 $\pm$ 0.015 & 0.683 $\pm$ 0.047 & 0.711 $\pm$ 0.026  \\
1D-CNN    &  0.805 $\pm$ 0.017   &  0.794 $\pm$ 0.007      &  0.740 $\pm$ 0.049 & 0.768 $\pm$ 0.054 & 0.750 $\pm$ 0.007 
\end{tabular}}
\caption{Original raw data without pre-processing.}
\label{exp-1}
\end{table}
\FloatBarrier

In this case, SVM provided the best performance in terms of balanced accuracy, followed by 1D-CNN indicating the ability of the convolutional neural network to extract features from the samples.

\begin{table}[htbp]
\centering
\resizebox{\textwidth}{!}{\begin{tabular}{l|l|l|l|l|l}
\hline
Algorithm & ACC      & BACC     & Recall & Precision & F-Score \\ \hline
PLS-DA    &  0.785 $\pm$ 0.006   &  0.747 $\pm$ 0.006      &  0.566 $\pm$ 0.010 & 0.835 $\pm$ 0.016 & 0.675 $\pm$ 0.008   \\
SVM       &  \textbf{0.822 $\pm$ 0.012}   &  \textbf{0.795 $\pm$ 0.017}      &  0.667 $\pm$ 0.042 & \textbf{0.849 $\pm$ 0.010} & \textbf{0.746 $\pm$ 0.025}  \\
XGBoost   &  0.767 $\pm$ 0.025   &  0.776 $\pm$ 0.016      &  \textbf{0.818 $\pm$ 0.030} & 0.672 $\pm$ 0.044 & 0.736 $\pm$ 0.015 \\
CatBoost  &  0.791 $\pm$ 0.010   &  0.771 $\pm$ 0.014      &  0.675 $\pm$ 0.039 & 0.768 $\pm$ 0.024 & 0.718 $\pm$ 0.020 \\
LightGBM  &  0.794 $\pm$ 0.023   &  0.788 $\pm$ 0.013      &  0.761 $\pm$ 0.081 & 0.742 $\pm$ 0.068 & 0.744 $\pm$ 0.015  \\
1D-CNN    &  0.805 $\pm$ 0.013   &  0.782 $\pm$ 0.018      &  0.672 $\pm$ 0.052 & 0.805 $\pm$ 0.034 & 0.731 $\pm$ 0.025  
\end{tabular}}
\caption{Data pre-processed with SNV.}
\label{exp-2}
\end{table}
\FloatBarrier

Table \ref{exp-2} presents the results using SNV. A slight improvement can be noted in all algorithms. This kind of improvement is expected, as SNV is a standard technique for processing spectral data.

Despite the GBM algorithms achieving better results than in the previous experiment, SVM still achieves the best results in all metrics, with a BACC of $0.795$.

To finalize the evaluation stage of data pre-processing, Table \ref{exp-3} presents the results using pre-processing with SNV, in addition to the use of feature extraction.

\begin{table}[htbp]
\centering
\resizebox{\textwidth}{!}{\begin{tabular}{l|l|l|l|l|l}
\hline
Algorithm & ACC &  BACC        & Recall & Precision & F-Score \\ \hline
PLS-DA    &  0.812 $\pm$ 0.002   &  0.778 $\pm$ 0.003      &  0.613 $\pm$ 0.010 & \textbf{0.874 $\pm$ 0.011} & 0.721 $\pm$ 0.005   \\
SVM       &  0.791 $\pm$ 0.015   &  0.762 $\pm$ 0.014      &  0.626 $\pm$ 0.025 & 0.803 $\pm$ 0.041 & 0.703 $\pm$ 0.019  \\
XGBoost   &  0.798 $\pm$ 0.008   &  0.802 $\pm$ 0.005      &  \textbf{0.821 $\pm$ 0.025} & 0.713 $\pm$ 0.019 & 0.762 $\pm$ 0.006 \\
CatBoost  &  0.799 $\pm$ 0.010   &  0.777 $\pm$ 0.010      &  0.672 $\pm$ 0.021 & 0.788 $\pm$ 0.022 & 0.725 $\pm$ 0.013 \\
LightGBM  &  \textbf{0.820 $\pm$ 0.010}   &  \textbf{0.815 $\pm$ 0.004}      &  0.792 $\pm$ 0.034 & 0.764 $\pm$ 0.036 & \textbf{0.776 $\pm$ 0.004}  \\
1D-CNN    &  -            &  -            &  -          & - & -        
\end{tabular}}
\caption{Data pre-processed with SNV and feature extraction.}
\label{exp-3}
\end{table}
\FloatBarrier

Data processed with feature extraction provided the best results with the GBM models, especially the LightGBM algorithm  with balanced accuracy of $0.815$.

The 1D-CNN algorithm was not listed in Table \ref{exp-3}, as a manual feature extraction would not make sense as pre-processing since convolution layers act as implicit feature extractors in  neural networks.

\subsection{Results obtained using SMOTE}
For the creation of synthetic samples, only the initial training set was used, generating a balance between the classes, while the validation and test base remains intact.

\begin{table}[htbp]
\centering
\resizebox{\textwidth}{!}{\begin{tabular}{l|l|l|l|l|l}
\hline
Algorithm & ACC &  BACC & Recall & Precision & F-Score \\ \hline
PLS-DA    &  0.723 $\pm$ 0.007   &  0.717 $\pm$ 0.009      &  0.685 $\pm$ 0.019 & 0.639 $\pm$ 0.007 & 0.662 $\pm$ 0.012   \\
SVM       &  \textbf{0.819 $\pm$ 0.013}   &  \textbf{0.812 $\pm$ 0.011}      &  0.784 $\pm$ 0.021 & \textbf{0.764 $\pm$ 0.025} & \textbf{0.773 $\pm$ 0.013}  \\
XGBoost   &  0.743 $\pm$ 0.020   &  0.756 $\pm$ 0.018      &  \textbf{0.818 $\pm$ 0.031} & 0.636 $\pm$ 0.024 & 0.715 $\pm$ 0.019 \\
CatBoost  &  0.757 $\pm$ 0.020   &  0.753 $\pm$ 0.020      &  0.735 $\pm$ 0.021 & 0.678 $\pm$ 0.026 & 0.705 $\pm$ 0.022 \\
LightGBM  &  0.751 $\pm$ 0.024   &  0.754 $\pm$ 0.031      &  0.771 $\pm$ 0.085 & 0.658 $\pm$ 0.027 & 0.708 $\pm$ 0.039  \\
1D-CNN    &  0.774 $\pm$ 0.026   &  0.774 $\pm$ 0.023      &  0.774 $\pm$ 0.051 & 0.695 $\pm$ 0.043 & 0.730 $\pm$ 0.026    
\end{tabular}}
\caption{Original data without pre-processing.}
\label{exp-4}
\end{table}

Table \ref{exp-4} presents the results using SMOTE without pre-processing. One can notice an improvement in the performance of some models, when compared with data in Table \ref{exp-1}. With a greater number of samples, the SVM becomes the best model among those tested with balanced accuracy of $0.812$.

\begin{table}[htbp]
\centering
\resizebox{\textwidth}{!}{\begin{tabular}{l|l|l|l|l|l}
\hline
Algorithm & ACC & BACC & Recall & Precision & F-Score \\ \hline
PLS-DA    &  0.743 $\pm$ 0.006   &  0.743 $\pm$ 0.007      &  0.745 $\pm$ 0.016 & 0.652 $\pm$ 0.007 & 0.696 $\pm$ 0.009   \\
SVM       &  \textbf{0.820 $\pm$ 0.008}   &  \textbf{0.813 $\pm$ 0.006}      &  \textbf{0.779 $\pm$ 0.014} & 0.768 $\pm$ 0.021 & \textbf{0.773 $\pm$ 0.007}  \\
XGBoost   &  0.778 $\pm$ 0.049   &  0.778 $\pm$ 0.035      &  0.779 $\pm$ 0.064 & 0.711 $\pm$ 0.084 & 0.737 $\pm$ 0.032 \\
CatBoost  &  0.808 $\pm$ 0.017   &  0.795 $\pm$ 0.020      &  0.730 $\pm$ 0.047 & 0.773 $\pm$ 0.029 & 0.750 $\pm$ 0.026 \\
LightGBM  &  0.756 $\pm$ 0.030   &  0.757 $\pm$ 0.021      &  0.763 $\pm$ 0.048 & 0.673 $\pm$ 0.049 & 0.713 $\pm$ 0.021  \\
1D-CNN    &  0.808 $\pm$ 0.012   &  0.788 $\pm$ 0.010      &  0.693 $\pm$ 0.031 & \textbf{0.798 $\pm$ 0.041} & 0.741 $\pm$ 0.013
\end{tabular}}
\caption{Data pre-processed with SNV.}
\label{exp-5}
\end{table}

Table \ref{exp-5} presents the results using SNV. The model that presented the best balanced accuracy was SVM with value of $0.813$.

For the last experiment using SMOTE, Table \ref{exp-6} lists the results using SNV and feature extraction with a value of BACC of $0.790$ provided by CatBoost and LightGBM and PLS-DA. 
The 1D-CNN are not applied in this experiment since convolutional layers extract features of the data implicitly.

\begin{table}[htbp]
\centering
\resizebox{\textwidth}{!}{\begin{tabular}{l|l|l|l|l|l}
\hline
Algorithm & ACC & BACC & Recall & Precision & F-Score \\ \hline
PLS-DA    &  0.798 $\pm$ 0.005   &  \textbf{0.790 $\pm$ 0.005}      &  0.753 $\pm$ 0.008 & 0.740 $\pm$ 0.010 & 0.747 $\pm$ 0.005   \\
SVM       &  0.789 $\pm$ 0.013   &  0.776 $\pm$ 0.016      &  0.717 $\pm$ 0.041 & 0.741 $\pm$ 0.019 & 0.728 $\pm$ 0.021  \\
XGBoost   &  0.784 $\pm$ 0.011   &  0.785 $\pm$ 0.009      &  \textbf{0.789 $\pm$ 0.027} & 0.704 $\pm$ 0.023 & 0.743 $\pm$ 0.010 \\
CatBoost  &  \textbf{0.804 $\pm$ 0.019}   &  \textbf{0.790 $\pm$ 0.018}      &  0.724 $\pm$ 0.025 & \textbf{0.768 $\pm$ 0.034} & 0.745 $\pm$ 0.022 \\
LightGBM  &  0.796 $\pm$ 0.009   &  0.790 $\pm$ 0.010      &  0.763 $\pm$ 0.025 & 0.732 $\pm$ 0.016 & \textbf{0.747 $\pm$ 0.012}  \\
1D-CNN    &  -           &  -            &  -          & - & - 
\end{tabular}}
\caption{Data Pre-processed with SNV, and with feature extraction.}
\label{exp-6}
\end{table}
\FloatBarrier

\subsection{Results obtained with GAN}

Using synthetic samples generated using GAN, table \ref{exp-7} presents the results for the case of no pre-processing is performed.

\begin{table}[htbp]
\centering
\resizebox{\textwidth}{!}{\begin{tabular}{l|l|l|l|l|l}
\hline
Algorithm & ACC & BACC & Recall & Precision & F-Score \\ \hline
PLS-DA    &  0.689 $\pm$ 0.014   &  0.675 $\pm$ 0.015           &  0.690 $\pm$ 0.014  & 0.689 $\pm$ 0.014 & 0.690 $\pm$ 0.001    \\
SVM       &  \textbf{0.818 $\pm$ 0.018}   &  0.787 $\pm$ 0.019           &  \textbf{0.818 $\pm$ 0.018}  & \textbf{0.827 $\pm$ 0.018} & \textbf{0.812 $\pm$ 0.019}   \\
XGBoost   &  0.771 $\pm$ 0.025   &  0.762 $\pm$ 0.025           &  0.771 $\pm$ 0.025  & 0.772 $\pm$ 0.024 & 0.771 $\pm$ 0.025  \\
CatBoost  &  0.788 $\pm$ 0.014   &  0.769 $\pm$ 0.016           &  0.788 $\pm$ 0.014  & 0.787 $\pm$ 0.014 & 0.785 $\pm$ 0.014  \\
LightGBM  &  0.787 $\pm$ 0.002   &  0.779 $\pm$ 0.019           &  0.788 $\pm$ 0.015  & 0.788 $\pm$ 0.017 & 0.788 $\pm$ 0.016   \\
1D-CNN    &  0.808 $\pm$ 0.014   &  \textbf{0.793 $\pm$ 0.018}           &  0.808 $\pm$ 0.014  & 0.808 $\pm$ 0.014 & 0.806 $\pm$ 0.014 
\end{tabular}}
\caption{Original spectral data, without pre-processing.}
\label{exp-7}
\end{table}

Considering the results of previous experiments with no-preprocessing the data, SVM and 1D-CNN algorithms provide the best results in terms of BACC from original data, while GBM algorithms provided inferior performance in terms of BACC without pre-processing.

Table \ref{exp-8} presents the result using SNV. One observes the performance of  SVM  in terms of BACC is 0.790. The results indicate in this experiment that SVM provided the  best performance for most of the metrics.

\begin{table}[htbp]
\centering
\resizebox{\textwidth}{!}{\begin{tabular}{l|l|l|l|l|l}
\hline
Algorithm & ACC & BACC & Recall & Precision & F-Score \\ \hline
PLS-DA    &  0.712 $\pm$ 0.015   &  0.699 $\pm$ 0.012            &  0.712 $\pm$ 0.015 & 0.712 $\pm$ 0.001 & 0.712 $\pm$ 0.014    \\
SVM       &  \textbf{0.816 $\pm$ 0.014}   &  \textbf{0.790 $\pm$ 0.019}            &  \textbf{0.816 $\pm$ 0.014} & \textbf{0.820 $\pm$ 0.013} & \textbf{0.811 $\pm$ 0.016}   \\
XGBoost   &  0.794 $\pm$ 0.030   &  0.778 $\pm$ 0.038            &  0.794 $\pm$ 0.030 & 0.795 $\pm$ 0.031 & 0.791 $\pm$ 0.033  \\
CatBoost  &  0.765 $\pm$ 0.009   &  0.739 $\pm$ 0.012            &  0.765 $\pm$ 0.009 & 0.766 $\pm$ 0.010 & 0.760 $\pm$ 0.010  \\
LightGBM  &  0.806 $\pm$ 0.023   &  0.790 $\pm$ 0.024            &  0.806 $\pm$ 0.023 & 0.805 $\pm$ 0.024 & 0.804 $\pm$ 0.023   \\
1D-CNN    &  0.806 $\pm$ 0.011   &  0.785 $\pm$ 0.012            &  0.806 $\pm$ 0.011 & 0.808 $\pm$ 0.013 & 0.803 $\pm$ 0.011
\end{tabular}}
\caption{Data pre-processed with SNV.}
\label{exp-8}
\end{table}

 Table \ref{exp-9} presents the results for data with SNV and feature extraction. In this experiment, LightGBM presents the best results among all the algorithms with BACC of $0.839$.  CatBoost also performs very competitive to LightGBM.

\begin{table}[htbp]
\centering
\resizebox{\textwidth}{!}{\begin{tabular}{l|l|l|l|l|l}
\hline
Algorithm & ACC &   BACC & Recall & Precision & F-Score \\ \hline
PLS-DA    &  0.771 $\pm$ 0.005   &  0.748 $\pm$ 0.006      &  0.771 $\pm$ 0.005 & 0.770 $\pm$ 0.005 & 0.767 $\pm$ 0.006    \\
SVM       &  0.780 $\pm$ 0.033   &  0.750 $\pm$ 0.031      &  0.780 $\pm$ 0.033 & 0.784 $\pm$ 0.036 & 0.773 $\pm$ 0.033   \\
XGBoost   &  0.798 $\pm$ 0.027   &  0.786 $\pm$ 0.023      &  0.798 $\pm$ 0.027 & 0.804 $\pm$ 0.031 & 0.797 $\pm$ 0.025  \\
CatBoost  &  0.845 $\pm$ 0.025   &  0.819 $\pm$ 0.027      &  0.845 $\pm$ 0.025 & 0.852 $\pm$ 0.028 & 0.840 $\pm$ 0.026  \\
LightGBM  &  \textbf{0.851 $\pm$ 0.017}   &  \textbf{0.839 $\pm$ 0.018}      &  \textbf{0.851 $\pm$ 0.017} & \textbf{0.852 $\pm$ 0.017} & \textbf{0.850 $\pm$ 0.017}   \\
1D-CNN    &  -    &  -             &  -     & -   & -  
\end{tabular}}
\caption{Data pre-processed with SNV, and with feature extraction.}
\label{exp-9}
\end{table}
\FloatBarrier

\subsection{Feature Importance with SHAP}
\begin{figure}[ht]
    \includegraphics[width=0.9\linewidth]{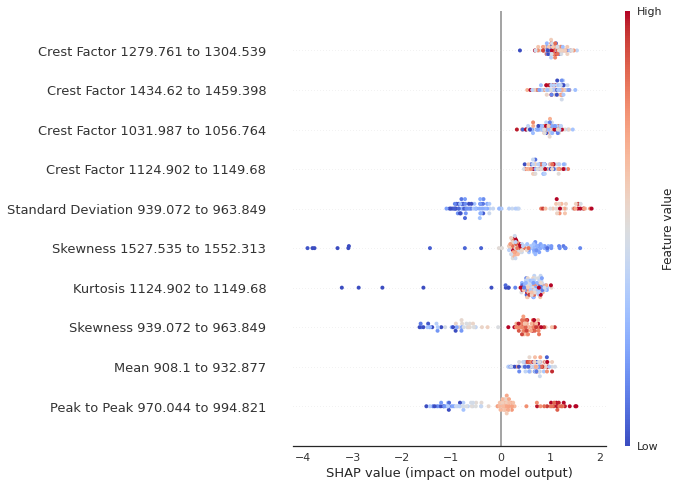}
    \caption{SHAP importance for each feature extracted from the subsequences of the test set. The $x$ axes indicates the calculated SHAP value on the model output, while the $y$ axes display each feature, sorted by its importance for the classifier. The color indicates the original feature value. }
    \label{shap_means}
\end{figure}

We applied the SHAP (SHapley Additive exPlanations) algorithm \citep{shap} to assess feature importance of the model and to get a visual representation of the statistical features of the subsequences. The SHAP  algorithm can be used to explain the output of any machine learning model. 
For the skin cancer classification model, it is possible to observe the average absorption value for each subsequence created and see which one is the most relevant to the model. In order to illustrate the SHAP applied to this case, Figure \ref{shap_means} shows the feature importance analysis only for the LightGBM model, which provides the best overall results with SNV, feature extraction and data augmentation with GAN.

\FloatBarrier

The application of SHAP for the LightGBM model provides information that some extracted features such as standard deviation in the wavelength from 939 nm to 963 nm, skewness from 939 nm to 963 nm and peak to peak from 970 nm to 994 nm are important features in the classification.

\subsection{Discussion}

In terms of overall performance obtained in the experiments, one notes that some algorithms perform better than others depending on the type of pre-processing used. SVM presents consistent performance results in all experiments, while PLS-DA presents slightly worse results without any type of pre-processing, and more competitive ones using extracted features and data augmentation.

GBM algorithms also perform better when feature extraction is used. They achieve the best performance amongst all the experiments using GAN to generate synthetic samples. In the experiments, Catboost and LightGBM results are very close, with a slightly advantage to LightGBM.

One notices that 1D-CNN present a better performance in reducing false negatives and false positives when the sample signal is pre-processed with SNV. Generally Neural Networks require a large amount of data in its training, which was not the case in this study. Regarding the pre-processing step, convolutional layers work in a way that features are extracted implicitly, and there is no need to extract them. This may represent an advantage of 1D-CNN.

In their work, \citet{Arajo2021} utilized Raman spectra and employed the LightGBM model to obtain the best results. It is worth noting that they worked with a dataset comprising 436 spectral samples, with 168 belonging to nevus and 268 to melanoma. However, the data they utilized for their experiments is not publicly available, thereby making it impossible to directly compare our results.

% \subsection{Deployment}

% Using the test samples, it is possible to analyse in detail the performance of the model in a production environment. Table \ref{deploy-table} presents the total of cancer and non-cancer predictions of the LightGBM model trained with SNV and feature extraction samples for each skin lesion present in the test set.

% \begin{table}[htbp]
% \centering
% \begin{tabular}{l|l|l}
% \hline
% Lesion    & Cancer   &  Non-Cancer \\ \hline
% ACK    &  2/34   &  32/34   \\
% SEK    &  0/19   &  19/19   \\
% NEV    &  0/6   &  6/6   \\
% MEL    &  0/1   &  1/1   \\
% CBC    &  24/31   &  7/31   \\
% CEC    &  4/7   &  3/7   \\
% \end{tabular}
% \caption{Analysis of each lesion classified by the best performing model LightGBM for cancer and non-cancer skin lesions.}
% \label{deploy-table}
% \end{table}

%  It is important to note that the class imbalance within our test set, where certain lesions, such as Melanoma, have a limited number of samples. This poses a challenge in automated diagnosis, both in general and when utilizing NIR spectral data. Nonetheless, our model has shown promising results.

\citet{krohling2021} proposed an approach that combines clinical images, clinical information of the lesion, and patients' demographics. They trained a Convolutional Neural Network (CNN) using this aggregated information, achieving an average balanced accuracy of 0.85. Our best-performing LightGBM model investigated in this work using NIR spectral samples reveals a competitive performance in skin cancer detection. It is worth mentioning that our results show promise  with the yet limited size of our collected dataset. 

Our goal is to publish this dataset as \textit{Data in Brief} and the source code is available under request from authors during submission. After publication, the source code used in our experiments will be available on GitHub.

%%%%%%%%%%%%%%%%%%%%%%%%%%%%%%%%
%%%%%%%%%%%%%%%%%%%%%%%%%%%%%%%%
\section{Conclusion}

 In this paper, we investigate the suitability of NIR spectral data on skin cancer automated diagnosis using machine and deep learning algorithms aiming to develop a triage tool. Firstly, we introduced a new dataset named NIR-SC-UFES composed of spectral samples collected via a portable Micronir spectrometer in the wavelength  range of 900 nm to 1700 nm of the most common skin lesions \textit{in vivo} grouped as 1) cancer: melanoma (MEL), basal cellular carcinoma (BCC) and squamous cellular carcinoma (SCC); and 2) non-cancer: nevus (NEV), actinic keratosis (ACK) and seborrheic keratosis (SEK). As  far as we know, this is the first time such a dataset is collected since there is no public dataset available of this sort.  Secondly, since the dataset is imbalanced, we use a SMOTE-based data augmentation and GAN-based oversampling to generate synthetic data. Next, we apply the most powerful machine learning algorithms XGBoost, CatBoost, LightGBM, 1D-convolutional neural networks (1D-CNN), as well as standard algorithms as SVM and PLS-DA to our dataset. Experimental results indicate the best performance obtained by LightGBM with pre-processing using standard normal variate  (SNV), feature extraction and data augmentation with Generative Adversarial Networks (GAN)  providing values of 0.839 for balanced accuracy, 0.851 for recall, 0.852 for precision, and 0.850 for F-score. Through the application of SHAP, we have discovered that the most crucial range of the spectra lies within the interval of 939.072 nm to 994.821 nm. Our dataset NIR-SC-UFES, has a relatively small number of spectral samples for certain skin lesions, particularly melanoma. Indeed, this limitation exists and is currently being addressed. We are exploring additional sources of information to further enhance our results in order to assist doctors in the screening process.

%%%%%%%%%%%%%%%%%%%%%%%%%%%%%%%%

% \appendix

% \section{Sample Appendix Section}
% \label{sec:sample:appendix}
% Lorem ipsum dolor sit amet, consectetur adipiscing elit, sed do eiusmod tempor section \ref{sec:sample1} incididunt ut labore et dolore magna aliqua. Ut enim ad minim veniam, quis nostrud exercitation ullamco laboris nisi ut aliquip ex ea commodo consequat. Duis aute irure dolor in reprehenderit in voluptate velit esse cillum dolore eu fugiat nulla pariatur. Excepteur sint occaecat cupidatat non proident, sunt in culpa qui officia deserunt mollit anim id est laborum.

\section{Acknowledgments}
 The authors thank Isadora Tavares Nascimento and Isabella Rezende for the support during the time collecting the dataset NIR-SC-UFES at PAD-UFES. R.A. Krohling thanks the Brazilian research agency Conselho Nacional de Desenvolvimento Científico e Tecnológico (CNPq), Brazil - grant no. 304688/2021-5 and the Fundação de Amparo à Pesquisa e Inovação do Espírito Santo (FAPES), Brazil – grant no. 21/2022. P.F. Filgueiras thanks the support of CNCPq (409700/2022-3, and 305459/2020-1) and Fapes (442/2021, 691/2022 P: 2022-2DRM4, 1036/2022 P: 2022-VZ8G9 e 343/2023 P 2023-6SJG7).
 
\bibliographystyle{elsarticle-harv} 
\bibliography{bibliography}

\begin{thebibliography}{34}
\expandafter\ifx\csname natexlab\endcsname\relax\def\natexlab#1{#1}\fi
\providecommand{\url}[1]{\texttt{#1}}
\providecommand{\href}[2]{#2}
\providecommand{\path}[1]{#1}
\providecommand{\DOIprefix}{doi:}
\providecommand{\ArXivprefix}{arXiv:}
\providecommand{\URLprefix}{URL: }
\providecommand{\Pubmedprefix}{pmid:}
\providecommand{\doi}[1]{\href{http://dx.doi.org/#1}{\path{#1}}}
\providecommand{\Pubmed}[1]{\href{pmid:#1}{\path{#1}}}
\providecommand{\bibinfo}[2]{#2}
\ifx\xfnm\relax \def\xfnm[#1]{\unskip,\space#1}\fi
%Type = Inproceedings
\bibitem[{Abayomi-Alli et~al.(2020)Abayomi-Alli, Dama{\v{s}}evi{\v{c}}ius,
  Wieczorek and Wo{\'z}niak}]{alli2020}
\bibinfo{author}{Abayomi-Alli, O.O.},
  \bibinfo{author}{Dama{\v{s}}evi{\v{c}}ius, R.}, \bibinfo{author}{Wieczorek,
  M.}, \bibinfo{author}{Wo{\'z}niak, M.}, \bibinfo{year}{2020}.
\newblock \bibinfo{title}{Data augmentation using principal component
  resampling for image recognition by deep learning}, in:
  \bibinfo{booktitle}{Artificial Intelligence and Soft Computing: 19th
  International Conference, ICAISC 2020, Zakopane, Poland, October 12-14, 2020,
  Proceedings, Part II 19}, \bibinfo{organization}{Springer}. pp.
  \bibinfo{pages}{39--48}.
%Type = Inproceedings
\bibitem[{Akiba et~al.(2019)Akiba, Sano, Yanase, Ohta and Koyama}]{Akiba2019}
\bibinfo{author}{Akiba, T.}, \bibinfo{author}{Sano, S.},
  \bibinfo{author}{Yanase, T.}, \bibinfo{author}{Ohta, T.},
  \bibinfo{author}{Koyama, M.}, \bibinfo{year}{2019}.
\newblock \bibinfo{title}{Optuna: A next-generation hyperparameter optimization
  framework}, in: \bibinfo{booktitle}{Proceedings of the 25th {ACM} {SIGKDD}
  International Conference on Knowledge Discovery {\&} Data Mining},
  \bibinfo{publisher}{Association for Computing Machinery},
  \bibinfo{address}{New York, NY, USA}. p. \bibinfo{pages}{2623–2631}.
%Type = Article
\bibitem[{Ara{\'{u}}jo et~al.(2021)Ara{\'{u}}jo, Veloso, de~Oliveira~Filho,
  Giraud, Raniero, Ferreira and Bitar}]{Arajo2021}
\bibinfo{author}{Ara{\'{u}}jo, D.C.}, \bibinfo{author}{Veloso, A.A.},
  \bibinfo{author}{de~Oliveira~Filho, R.S.}, \bibinfo{author}{Giraud, M.N.},
  \bibinfo{author}{Raniero, L.J.}, \bibinfo{author}{Ferreira, L.M.},
  \bibinfo{author}{Bitar, R.A.}, \bibinfo{year}{2021}.
\newblock \bibinfo{title}{Finding reduced {R}aman spectroscopy fingerprint of
  skin samples for melanoma diagnosis through machine learning}.
\newblock \bibinfo{journal}{Artificial Intelligence in Medicine}
  \bibinfo{volume}{120}, \bibinfo{pages}{102161}.
%Type = Article
\bibitem[{Barnes et~al.(1989)Barnes, Dhanoa and Lister}]{snv}
\bibinfo{author}{Barnes, R.J.}, \bibinfo{author}{Dhanoa, M.S.},
  \bibinfo{author}{Lister, S.J.}, \bibinfo{year}{1989}.
\newblock \bibinfo{title}{Standard normal variate transformation and
  de-trending of near-infrared diffuse reflectance spectra}.
\newblock \bibinfo{journal}{Applied Spectroscopy} \bibinfo{volume}{43},
  \bibinfo{pages}{772--777}.
%Type = Article
\bibitem[{Bergstra et~al.(2011)Bergstra, Bardenet, Bengio and
  K{\'e}gl}]{Bergstra2011}
\bibinfo{author}{Bergstra, J.}, \bibinfo{author}{Bardenet, R.},
  \bibinfo{author}{Bengio, Y.}, \bibinfo{author}{K{\'e}gl, B.},
  \bibinfo{year}{2011}.
\newblock \bibinfo{title}{Algorithms for hyper-parameter optimization}.
\newblock \bibinfo{journal}{Advances in Neural Information Processing Systems}
  \bibinfo{volume}{24}.
%Type = Inproceedings
\bibitem[{Boser et~al.(1992)Boser, Guyon and Vapnik}]{svm}
\bibinfo{author}{Boser, B.E.}, \bibinfo{author}{Guyon, I.M.},
  \bibinfo{author}{Vapnik, V.N.}, \bibinfo{year}{1992}.
\newblock \bibinfo{title}{A training algorithm for optimal margin classifiers},
  in: \bibinfo{booktitle}{Proceedings of the 5th Annual Workshop on
  Computational Learning Theory}, pp. \bibinfo{pages}{144--152}.
%Type = Article
\bibitem[{Brereton(2015)}]{BRERETON201590}
\bibinfo{author}{Brereton, R.G.}, \bibinfo{year}{2015}.
\newblock \bibinfo{title}{Pattern recognition in chemometrics}.
\newblock \bibinfo{journal}{Chemometrics and Intelligent Laboratory Systems}
  \bibinfo{volume}{149}, \bibinfo{pages}{90--96}.
%Type = Article
\bibitem[{Chawla et~al.(2002)Chawla, Bowyer, Hall and Kegelmeyer}]{smote}
\bibinfo{author}{Chawla, N.V.}, \bibinfo{author}{Bowyer, K.W.},
  \bibinfo{author}{Hall, L.O.}, \bibinfo{author}{Kegelmeyer, W.P.},
  \bibinfo{year}{2002}.
\newblock \bibinfo{title}{{SMOTE}: Synthetic minority over-sampling technique}.
\newblock \bibinfo{journal}{Journal of Artificial Intelligence Research}
  \bibinfo{volume}{16}, \bibinfo{pages}{321--357}.
%Type = Article
\bibitem[{Chen and Guestrin(2016)}]{Chen2016}
\bibinfo{author}{Chen, T.}, \bibinfo{author}{Guestrin, C.E.},
  \bibinfo{year}{2016}.
\newblock \bibinfo{title}{Xgboost: A scalable tree boosting system}.
\newblock \bibinfo{journal}{Knowledge Disc. Data Mining} ,
  \bibinfo{pages}{785–794}.
%Type = Article
\bibitem[{Friedman(2002)}]{Friedman2002}
\bibinfo{author}{Friedman, J.H.}, \bibinfo{year}{2002}.
\newblock \bibinfo{title}{Stochastic gradient boosting}.
\newblock \bibinfo{journal}{Computational Statistics \& Data Analysis}
  \bibinfo{volume}{38}, \bibinfo{pages}{367–378}.
%Type = Article
\bibitem[{Gniadecka et~al.(2004)Gniadecka, Philipsen, Wessel, Gniadecki, Wulf,
  Sigurdsson, Nielsen, Christensen, Hercogova, Rossen, Thomsen and
  Hansen}]{Gniadecka2004}
\bibinfo{author}{Gniadecka, M.}, \bibinfo{author}{Philipsen, P.A.},
  \bibinfo{author}{Wessel, S.}, \bibinfo{author}{Gniadecki, R.},
  \bibinfo{author}{Wulf, H.C.}, \bibinfo{author}{Sigurdsson, S.},
  \bibinfo{author}{Nielsen, O.F.}, \bibinfo{author}{Christensen, D.H.},
  \bibinfo{author}{Hercogova, J.}, \bibinfo{author}{Rossen, K.},
  \bibinfo{author}{Thomsen, H.K.}, \bibinfo{author}{Hansen, L.K.},
  \bibinfo{year}{2004}.
\newblock \bibinfo{title}{Melanoma diagnosis by {R}aman spectroscopy and neural
  networks: Structure alterations in proteins and lipids in intact cancer
  tissue}.
\newblock \bibinfo{journal}{Journal of Investigative Dermatology}
  \bibinfo{volume}{122}, \bibinfo{pages}{443--449}.
%Type = Book
\bibitem[{Goodfellow et~al.(2016)Goodfellow, Bengio and Courville}]{goodfellow}
\bibinfo{author}{Goodfellow, I.}, \bibinfo{author}{Bengio, Y.},
  \bibinfo{author}{Courville, A.}, \bibinfo{year}{2016}.
\newblock \bibinfo{title}{Deep Learning}.
\newblock \bibinfo{publisher}{MIT Press}.
\newblock \bibinfo{note}{\url{http://www.deeplearningbook.org}}.
%Type = Article
\bibitem[{Goodfellow et~al.(2014)Goodfellow, Pouget-Abadie, Mirza, Xu,
  Warde-Farley, Ozair, Courville and Bengio}]{goodfellow2014generative}
\bibinfo{author}{Goodfellow, I.}, \bibinfo{author}{Pouget-Abadie, J.},
  \bibinfo{author}{Mirza, M.}, \bibinfo{author}{Xu, B.},
  \bibinfo{author}{Warde-Farley, D.}, \bibinfo{author}{Ozair, S.},
  \bibinfo{author}{Courville, A.}, \bibinfo{author}{Bengio, Y.},
  \bibinfo{year}{2014}.
\newblock \bibinfo{title}{Generative adversarial nets}.
\newblock \bibinfo{journal}{Advances in Neural Information Processing Systems}
  \bibinfo{volume}{27}.
%Type = Article
\bibitem[{Gottfries et~al.(1995)Gottfries, K, A and C.}]{Gottfries1995}
\bibinfo{author}{Gottfries, J.}, \bibinfo{author}{K, B.}, \bibinfo{author}{A,
  W.}, \bibinfo{author}{C., G.}, \bibinfo{year}{1995}.
\newblock \bibinfo{title}{Diagnosis of dementias using partial least squares
  discriminant analysis. dementia geriatric cognit disorders.}
\newblock \bibinfo{journal}{Journal of Big Data} \bibinfo{volume}{6(2)},
  \bibinfo{pages}{83–8}.
%Type = Article
\bibitem[{Hancock et~al.(2020)Hancock, T., Khosgoftaar and M.}]{Hancock2020}
\bibinfo{author}{Hancock}, \bibinfo{author}{T., J.},
  \bibinfo{author}{Khosgoftaar}, \bibinfo{author}{M., T.},
  \bibinfo{year}{2020}.
\newblock \bibinfo{title}{Catboost for big data: an interdisciplinary review.}
\newblock \bibinfo{journal}{Journal of Big Data} \bibinfo{volume}{7},
  \bibinfo{pages}{2196--1115}.
%Type = Article
\bibitem[{Ke et~al.(2017)Ke, Meng, Finley, Wang, Chen, Ma, Ye and Liu}]{Ke2017}
\bibinfo{author}{Ke, G.}, \bibinfo{author}{Meng, Q.}, \bibinfo{author}{Finley,
  T.}, \bibinfo{author}{Wang, T.}, \bibinfo{author}{Chen, W.},
  \bibinfo{author}{Ma, W.}, \bibinfo{author}{Ye, Q.}, \bibinfo{author}{Liu,
  T.Y.}, \bibinfo{year}{2017}.
\newblock \bibinfo{title}{{LightGBM}: A highly efficient gradient boosting
  decision tree}.
\newblock \bibinfo{journal}{Advances in Neural Information Processing Systems}
  \bibinfo{volume}{30}.
%Type = Article
\bibitem[{Krohling et~al.(2021)Krohling, Castro, Pacheco and
  Krohling}]{krohling2021}
\bibinfo{author}{Krohling, B.}, \bibinfo{author}{Castro, P.B.},
  \bibinfo{author}{Pacheco, A.G.}, \bibinfo{author}{Krohling, R.A.},
  \bibinfo{year}{2021}.
\newblock \bibinfo{title}{A smartphone based application for skin cancer
  classification using deep learning with clinical images and lesion
  information}.
\newblock \bibinfo{journal}{arXiv preprint arXiv:2104.14353} .
%Type = Article
\bibitem[{Liu et~al.(2017)Liu, Osadchy, Ashton, Foster, Solomon and
  Gibson}]{Liu2017}
\bibinfo{author}{Liu, J.}, \bibinfo{author}{Osadchy, M.},
  \bibinfo{author}{Ashton, L.}, \bibinfo{author}{Foster, M.},
  \bibinfo{author}{Solomon, C.J.}, \bibinfo{author}{Gibson, S.J.},
  \bibinfo{year}{2017}.
\newblock \bibinfo{title}{Deep convolutional neural networks for {R}aman
  spectrum recognition: a unified solution}.
\newblock \bibinfo{journal}{The Analyst} \bibinfo{volume}{142},
  \bibinfo{pages}{4067--4074}.
%Type = Article
\bibitem[{Lundberg and Lee(2017)}]{shap}
\bibinfo{author}{Lundberg, S.M.}, \bibinfo{author}{Lee, S.I.},
  \bibinfo{year}{2017}.
\newblock \bibinfo{title}{A unified approach to interpreting model
  predictions}.
\newblock \bibinfo{journal}{Advances in Neural Information Processing Systems
  30 (NIP 2017)} , \bibinfo{pages}{4765–4774}.
%Type = Article
\bibitem[{Malek et~al.(2017)Malek, Melgani and Bazi}]{Malek2017}
\bibinfo{author}{Malek, S.}, \bibinfo{author}{Melgani, F.},
  \bibinfo{author}{Bazi, Y.}, \bibinfo{year}{2017}.
\newblock \bibinfo{title}{One-dimensional convolutional neural networks for
  spectroscopic signal regression}.
\newblock \bibinfo{journal}{Journal of Chemometrics} \bibinfo{volume}{32},
  \bibinfo{pages}{e2977}.
%Type = Article
\bibitem[{McIntosh et~al.(2002)McIntosh, Jackson, Mantsch, Mansfield, Crowson
  and Toole}]{McIntosh2002}
\bibinfo{author}{McIntosh, L.M.}, \bibinfo{author}{Jackson, M.},
  \bibinfo{author}{Mantsch, H.H.}, \bibinfo{author}{Mansfield, J.R.},
  \bibinfo{author}{Crowson, A.}, \bibinfo{author}{Toole, J.W.},
  \bibinfo{year}{2002}.
\newblock \bibinfo{title}{Near-infrared spectroscopy for dermatological
  applications}.
\newblock \bibinfo{journal}{Vibrational Spectroscopy} \bibinfo{volume}{28},
  \bibinfo{pages}{53--58}.
%Type = Article
\bibitem[{Morais et~al.(2020)Morais, Lima, Singh and Martin}]{Morais2020}
\bibinfo{author}{Morais, C.L.M.}, \bibinfo{author}{Lima, K.M.G.},
  \bibinfo{author}{Singh, M.}, \bibinfo{author}{Martin, F.L.},
  \bibinfo{year}{2020}.
\newblock \bibinfo{title}{Tutorial: multivariate classification for vibrational
  spectroscopy in biological samples}.
\newblock \bibinfo{journal}{Nature Protocols} \bibinfo{volume}{15},
  \bibinfo{pages}{2143--2162}.
%Type = Article
\bibitem[{Nair and Hinton(2010)}]{relu}
\bibinfo{author}{Nair, V.}, \bibinfo{author}{Hinton, G.E.},
  \bibinfo{year}{2010}.
\newblock \bibinfo{title}{Rectified linear units improve restricted {B}oltzmann
  machines}.
\newblock \bibinfo{journal}{International Conference on Machine Learning} ,
  \bibinfo{pages}{807–814}.
%Type = Article
\bibitem[{Ng et~al.(2020)Ng, Minasny, de~Sousa~Mendes and
  Dematt{\^{e}}}]{Ng2020}
\bibinfo{author}{Ng, W.}, \bibinfo{author}{Minasny, B.},
  \bibinfo{author}{de~Sousa~Mendes, W.}, \bibinfo{author}{Dematt{\^{e}},
  J.A.M.}, \bibinfo{year}{2020}.
\newblock \bibinfo{title}{The influence of training sample size on the accuracy
  of deep learning models for the prediction of soil properties with
  near-infrared spectroscopy data}.
\newblock \bibinfo{journal}{{SOIL}} \bibinfo{volume}{6},
  \bibinfo{pages}{565--578}.
%Type = Article
\bibitem[{Pacheco and Krohling(2020)}]{Pacheco2019}
\bibinfo{author}{Pacheco, A.G.}, \bibinfo{author}{Krohling, R.A.},
  \bibinfo{year}{2020}.
\newblock \bibinfo{title}{The impact of patient clinical information on
  automated skin cancer detection}.
\newblock \bibinfo{journal}{Computers in Biology and Medicine}
  \bibinfo{volume}{116}, \bibinfo{pages}{103545}.
%Type = Article
\bibitem[{Pacheco et~al.(2020)Pacheco, Lima, R, Salom{\~a}o, Krohling, Biral,
  de~Angelo, Alves~Jr, Esgario, Simora, C, Castro et~al.}]{Pacheco2020}
\bibinfo{author}{Pacheco, A.G.}, \bibinfo{author}{Lima}, \bibinfo{author}{R,
  G.}, \bibinfo{author}{Salom{\~a}o, A.S.}, \bibinfo{author}{Krohling, B.},
  \bibinfo{author}{Biral, I.P.}, \bibinfo{author}{de~Angelo, G.G.},
  \bibinfo{author}{Alves~Jr, F.C.}, \bibinfo{author}{Esgario, J.G.},
  \bibinfo{author}{Simora}, \bibinfo{author}{C, A.}, \bibinfo{author}{Castro,
  P.B.}, et~al., \bibinfo{year}{2020}.
\newblock \bibinfo{title}{{PAD-UFES-20}: a skin lesion dataset composed of
  patient data and clinical images collected from smartphones}.
\newblock \bibinfo{journal}{Data in Brief} \bibinfo{volume}{32},
  \bibinfo{pages}{106221}.
%Type = Article
\bibitem[{Pavlou and Kourkoumelis(2022)}]{PAVLOU2022104634}
\bibinfo{author}{Pavlou, E.}, \bibinfo{author}{Kourkoumelis, N.},
  \bibinfo{year}{2022}.
\newblock \bibinfo{title}{Deep adversarial data augmentation for biomedical
  spectroscopy: Application to modelling raman spectra of bone}.
\newblock \bibinfo{journal}{Chemometrics and Intelligent Laboratory Systems}
  \bibinfo{volume}{228}, \bibinfo{pages}{104634}.
%Type = Misc
\bibitem[{Prokhorenkova et~al.(2019)Prokhorenkova, Gusev, Vorobev, Dorogush and
  Gulin}]{Prokhorenkova2018}
\bibinfo{author}{Prokhorenkova, L.}, \bibinfo{author}{Gusev, G.},
  \bibinfo{author}{Vorobev, A.}, \bibinfo{author}{Dorogush, A.V.},
  \bibinfo{author}{Gulin, A.}, \bibinfo{year}{2019}.
\newblock \bibinfo{title}{Catboost: unbiased boosting with categorical
  features}.
\newblock \href{http://arxiv.org/abs/1706.09516}{{\tt arXiv:1706.09516}}.
%Type = Article
\bibitem[{Ramirez et~al.(2021)Ramirez, Greenop, Ashton and ur~Rehman}]{carlosa}
\bibinfo{author}{Ramirez, C.A.M.}, \bibinfo{author}{Greenop, M.},
  \bibinfo{author}{Ashton, L.}, \bibinfo{author}{ur~Rehman, I.},
  \bibinfo{year}{2021}.
\newblock \bibinfo{title}{Applications of machine learning in spectroscopy}.
\newblock \bibinfo{journal}{Applied Spectroscopy Reviews} \bibinfo{volume}{56},
  \bibinfo{pages}{733--763}.
%Type = Article
\bibitem[{Rumelhart et~al.(1986)Rumelhart, Hinton and Williams}]{nn}
\bibinfo{author}{Rumelhart, D.E.}, \bibinfo{author}{Hinton, G.E.},
  \bibinfo{author}{Williams, R.J.}, \bibinfo{year}{1986}.
\newblock \bibinfo{title}{Learning representations by back-propagating errors}.
\newblock \bibinfo{journal}{Nature} \bibinfo{volume}{323},
  \bibinfo{pages}{533--536}.
%Type = Inproceedings
\bibitem[{Smulko et~al.(2015)Smulko, Wrobel and Barman}]{Smulko2015}
\bibinfo{author}{Smulko, J.}, \bibinfo{author}{Wrobel, M.S.},
  \bibinfo{author}{Barman, I.}, \bibinfo{year}{2015}.
\newblock \bibinfo{title}{Noise in biological {R}aman spectroscopy}, in:
  \bibinfo{booktitle}{International Conference on Noise and Fluctuations
  ({ICNF})}, \bibinfo{publisher}{{IEEE}}. pp. \bibinfo{pages}{1--6}.
%Type = Book
\bibitem[{Stuart(2004)}]{stuart2005}
\bibinfo{author}{Stuart, B.H.}, \bibinfo{year}{2004}.
\newblock \bibinfo{title}{Infrared spectroscopy: fundamentals and
  applications}.
\newblock \bibinfo{publisher}{John Wiley \& Sons}.
%Type = Article
\bibitem[{Yuanyuan and Zhibin(2018)}]{Yuanyuan2018}
\bibinfo{author}{Yuanyuan, C.}, \bibinfo{author}{Zhibin, W.},
  \bibinfo{year}{2018}.
\newblock \bibinfo{title}{Quantitative analysis modeling of infrared
  spectroscopy based on ensemble convolutional neural networks}.
\newblock \bibinfo{journal}{Chemometrics and Intelligent Laboratory Systems}
  \bibinfo{volume}{181}, \bibinfo{pages}{1--10}.
%Type = Article
\bibitem[{Zeng et~al.(2021)Zeng, Wang, Xia, Li and Qu}]{Zeng_2021}
\bibinfo{author}{Zeng, W.}, \bibinfo{author}{Wang, Q.}, \bibinfo{author}{Xia,
  Z.}, \bibinfo{author}{Li, Z.}, \bibinfo{author}{Qu, H.},
  \bibinfo{year}{2021}.
\newblock \bibinfo{title}{Application of {XGBoost} algorithm in the detection
  of {SARS}-{CoV}-2 using {R}aman spectroscopy}.
\newblock \bibinfo{journal}{Journal of Physics: Conference Series}
  \bibinfo{volume}{1775}, \bibinfo{pages}{012007}.

\end{thebibliography}





% \end{thebibliography}
\end{document}